\documentclass[11pt]{article}

\usepackage[preprint]{acl}

\usepackage{times}
\usepackage{latexsym}

\usepackage[T1]{fontenc}

\usepackage[utf8]{inputenc}

\usepackage{microtype}

\usepackage{inconsolata}

\usepackage{graphicx}
\usepackage{dblfloatfix}
\usepackage{amsmath}
\usepackage{amssymb}
\usepackage{booktabs}
\usepackage{colortbl}
\usepackage{multirow}
\usepackage{array}
\usepackage{algorithm}
\usepackage{algpseudocode}
\usepackage[most]{tcolorbox}
\usepackage{tikz}
\usetikzlibrary{arrows.meta,positioning,calc}
\newcommand{\bestmetric}[1]{\textbf{#1}}
\newcommand{\secondmetric}[1]{#1}
\newcommand{\upmark}{\textsuperscript{\(\scriptstyle\uparrow\)}}
\newcommand{\downmark}{\textsuperscript{\(\scriptstyle\downarrow\)}}

\title{Bayesian-Agent: Posterior-Guided Skill Evolution Across LLM Agent Harnesses}

\author{
\bf Xiaojun Wu$^{\ast\ 1,2}$,
Cehao Yang$^{\ast\ 1,2}$,
Honghao Liu$^{\ast\ 1,2}$,
Xueyuan Lin$^{\ast\ 2}$, \\
\bf Wenjie Zhang$^{1}$,
Zhichao Shi$^{1}$,
Xuhui Jiang$^{1,3}$,  \\
\bf Chengjin Xu$^{1,3}$,
Jia Li$^{\dagger\ 2}$,
Jian Guo$^{\dagger\ 1}$
\\
\\
$^{1}$IDEA Research \\
$^{2}$The Hong Kong University of Science and Technology (Guangzhou) \\
$^{3}$DataArcTech Ltd.
}

\begin{document}
\maketitle
{
  \renewcommand{\thefootnote}%
    {\fnsymbol{footnote}}
  \footnotetext[1]{Equal Contribution}
  \footnotetext[2]{Corresponding Author}
}
\begin{abstract}
LLM agents increasingly rely on prompts, tools, memory, SOPs, skills, and harness feedback, yet current self-evolution pipelines often update these assets through heuristic reflection or raw success counts. Such updates are brittle when trajectories are sparse, expensive, and context-dependent. We introduce \textbf{Bayesian-Agent}, a native and cross-harness framework that treats reusable agent skills as Bayesian evidence objects. Bayesian-Agent records verified trajectories, maintains posterior beliefs over skill reliability and failure modes, and turns those beliefs into auditable skill actions and model-facing guardrails. This posterior view provides a finite-sample alternative to raw empirical-rate skill updates and frames prompt, context, and harness engineering as inference over the external decision environment. On RealFin-Bench, Bayesian skill evolution matches or improves the raw empirical-rate control and yields large gains on native runs. In the incremental mode, incremental repair improves SOP-Bench from 80\% to 95\%, Lifelong AgentBench from 90\% to 100\%, and RealFin-Bench from 45\% to 65\%. Backend and model-scaling ablations further show that posterior-guided repair can operate across BA native, MiniSWEAgent, and Claude Code, provided the harness produces verifiable task artifacts. 
The source code is available at \url{https://github.com/DataArcTech/Bayesian-Agent}.

\end{abstract}

\section{Introduction}
\label{sec:intro}

LLM agents increasingly depend on the environment around the model: prompts, tools, memory, retrieved context, standard operating procedures (SOPs), and verifier feedback \citep{react,toolformer,webarena,sweagent,openhands}. These assets determine what a fixed model can reliably accomplish, and skills make the external environment persistent across tasks \citep{memgpt,genericagent2026,voyager,skillsbench,skillweaver,sopagent}. The resulting problem is to maintain this environment as evidence accumulates, rather than to optimize model weights alone.

\begin{figure*}[t]
\centering
\includegraphics[width=\textwidth]{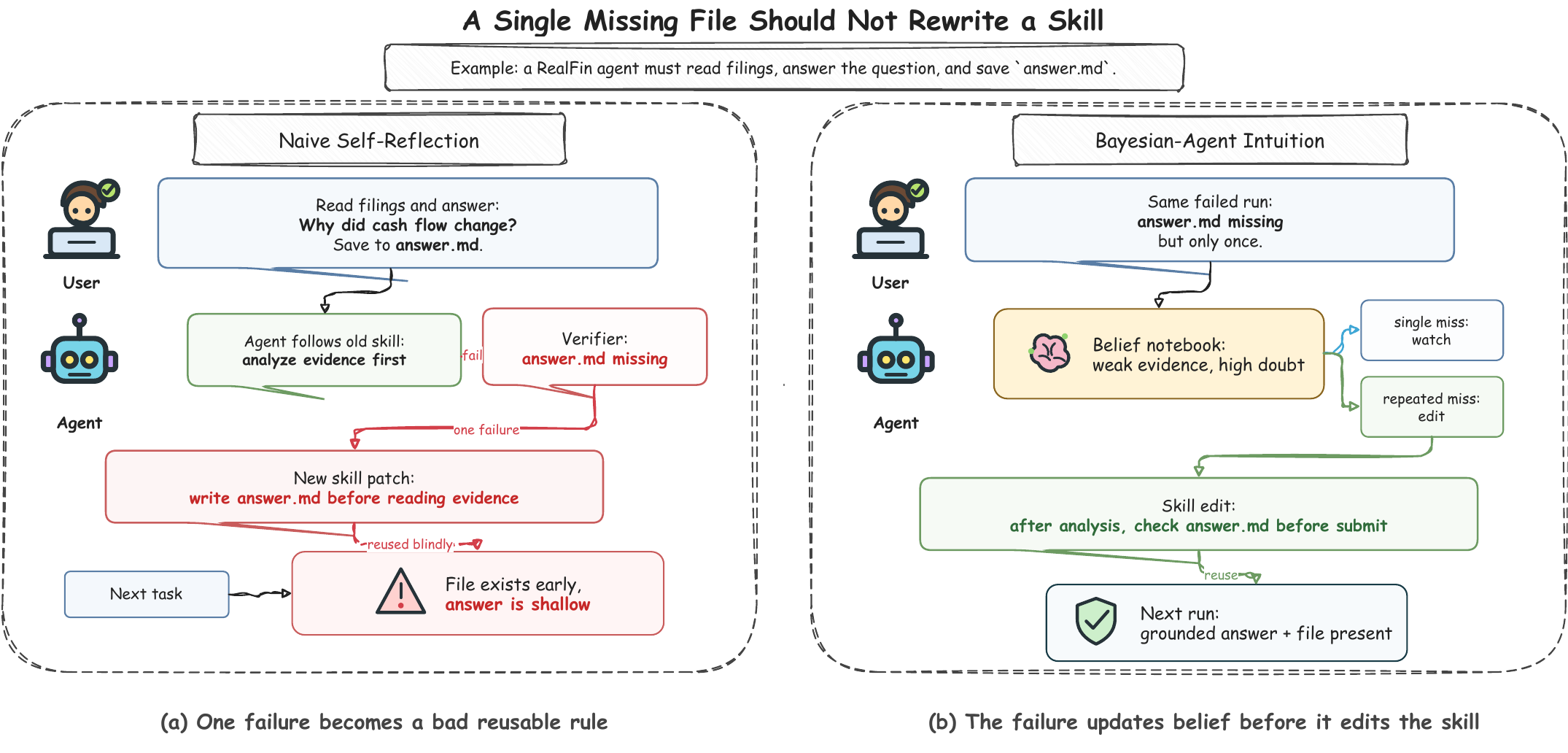}
\caption{Motivating example of sparse skill evidence. A single missing-output failure is too weak to justify a durable rule, whereas repeated verified failures support a targeted skill update.}
\label{fig:bayesian_agent_motivation}
\end{figure*}

Skill maintenance is difficult in the sparse online regime. A single failure may be accidental, context-specific, or the first indication of a reusable defect. A natural-language reflection loop can therefore turn a weak signal into a harmful rule, while a raw empirical rate can make an extreme decision from one or two observations. Figure~\ref{fig:bayesian_agent_motivation} illustrates this tension. Bayesian-Agent treats each reusable skill as a belief state with a prior, an evidence-conditioned posterior, and an uncertainty-aware action policy. This gives the harness a memory of evidence instead of restarting from a point estimate after every run.

Our central distinction is Bayesian rather than purely frequentist skill evolution. The empirical-rate controller uses observed successes divided by observations and has no explicit prior or uncertainty state. Bayesian-Agent combines a prior with verified trajectories to allocate expensive skill evaluations and choose whether to explore, patch, split, compress, or retire a skill. This is Bayesian optimization over a restricted harness-action space, not universal dominance: Section~\ref{sec:bayes_freq_theory} states when shrinkage lowers squared risk, and the experiments test its effect on completion. The inferred object is a persistent harness skill, not an individual answer.

We introduce \textbf{Bayesian-Agent}, a native and cross-harness skill-evolution layer. It converts verified trajectories into a compact feature-conditioned posterior, keeps the audit state separate from model-facing instructions, and renders executable guardrails and failure-mode patches for the next run. In \emph{full} mode the registry evolves online from an empty state; in \emph{incremental} mode an existing run supplies evidence and only failed tasks are repaired. The same evidence boundary supports a minimal native backend and adapters for GenericAgent, MiniSWEAgent, and Claude Code.

Our contributions are:
\begin{itemize}
  \item We formulate reusable agent skills as Bayesian evidence objects and give a finite-sample shrinkage analysis that makes the bias--variance trade-off explicit under sparse verified outcomes.
  \item We instantiate a unified Bayesian treatment of prompt, context, and harness engineering with an efficient categorical evidence model, posterior-guided skill actions, a native backend, and a cross-harness adapter boundary.
  \item We evaluate the framework on SOP-Bench, Lifelong AgentBench, and RealFin-Bench, including a Bayesian-vs.-raw-frequency control, full and incremental evolution, backend and model-scaling ablations, and skill-evolution case studies.
\end{itemize}

\section{Related Work}
\label{sec:rel_work}

\paragraph{LLM agents and harness engineering.}
ReAct and Toolformer established tool-mediated reasoning and action as a core agent pattern \citep{react,toolformer}. Subsequent systems expand the execution substrate with memory, multi-agent coordination, web interaction, and computer-use interfaces \citep{coala,generativeagents,metagpt,webarena,gaia,mind2web,sweagent,openhands}. GenericAgent further treats tools, memory, context, and self-evolution as engineering components of long-horizon execution \citep{genericagent2026}. Bayesian-Agent shares this harness-centric view, but models persistent skills and SOPs as evidence-bearing objects that can be maintained across tasks and execution substrates.

\paragraph{Self-evolving agents and skills.}
Reflexion and ExpeL turn trajectories into verbal feedback or experiential knowledge, while Voyager and related systems build reusable skill libraries \citep{reflexion,expel,voyager}. Recent work studies procedural skills, SOPs, computer-use skills, and evolving memory, and survey work positions skills as procedural infrastructure for LLM agents \citep{yang2026survey,skillsbench,skillweaver,sopagent,cuaskill,memskill}. SkillOpt and SkillGrad, together with textual-gradient methods such as TextGrad, optimize skill text in a richer train/selection/validation or train/test regime \citep{yang2026skillopt,wang2026skillgrad,yuksekgonul2024textgrad}. Bayesian-Agent addresses a different setting: few verified trajectories, no large skill-training pool, and online decisions about whether a skill should be patched, split, compressed, or retired.

\paragraph{Bayesian and evidence-guided agent optimization.}
Bayesian optimization allocates expensive evaluations by maintaining beliefs over uncertain outcomes \citep{snoek2012bo,shahriari2016bo,frazier2018bo,bergstra2011hyperopt}. Probabilistic modeling and calibration work provide complementary tools for representing uncertainty \citep{murphy2012ml,koller2009pgm,guo2017calibration,xiong2024uncertainty}. Recent LLM studies apply Bayesian reasoning to model decisions, confidence, or individual problem solutions, including BIRD and BayesAgent \citep{bird2025,bayesagent2026}. Our Bayesian object is different: a reusable skill in the external harness, whose posterior governs future context edits across tasks and harnesses rather than the solution path for one question.

\section{Method}
\label{sec:method}

\begin{figure*}[t]
\centering
\includegraphics[width=\textwidth]{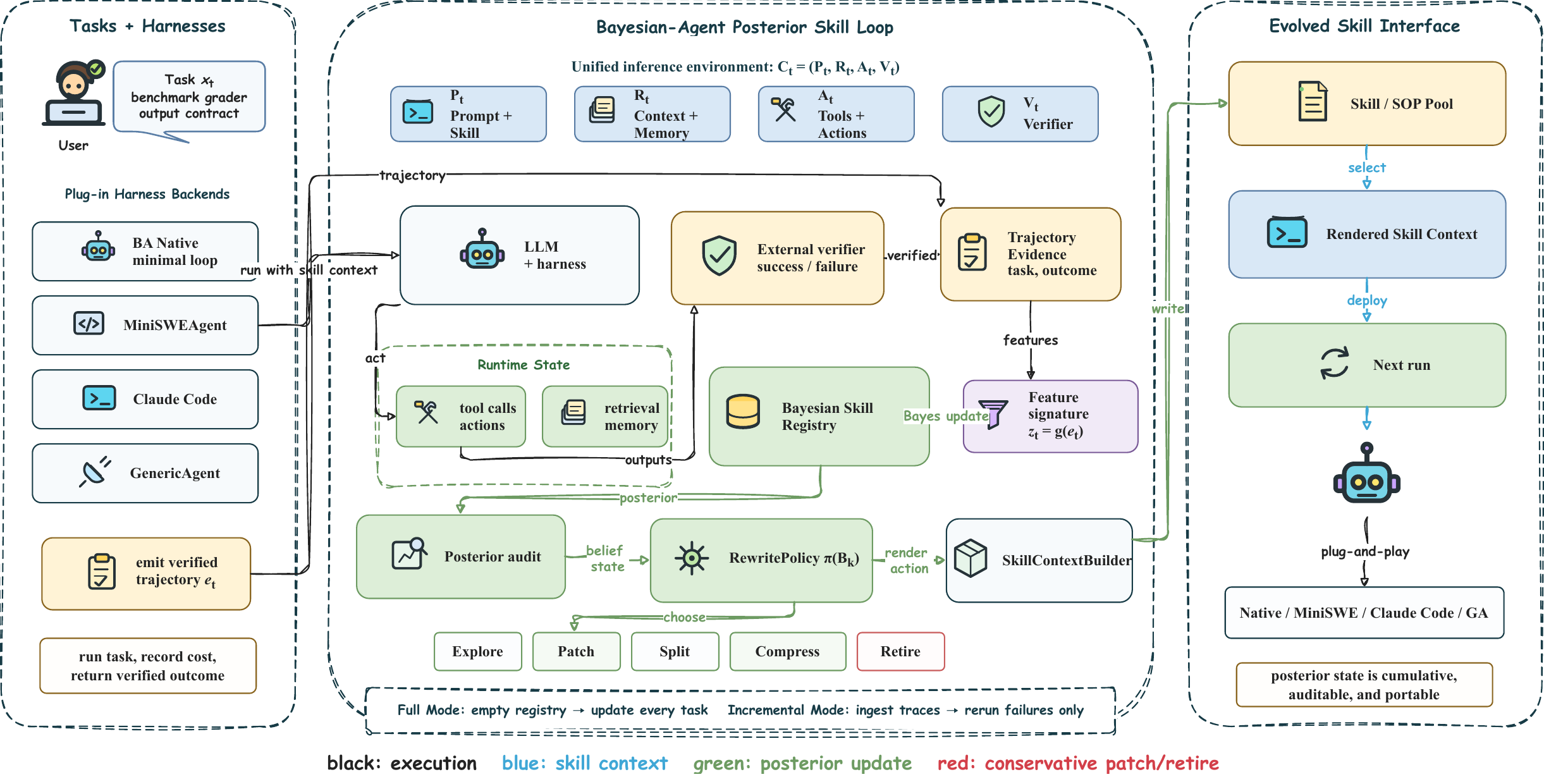}
\caption{Bayesian-Agent architecture. A native or external harness returns verified trajectories; the Bayesian layer extracts evidence, updates posterior skill beliefs, selects an auditable action, and renders executable skill context back to the harness.}
\label{fig:bayesian_agent_architecture}
\end{figure*}

Bayesian-Agent places a posterior evidence layer between an execution harness and the skill text shown to the model. Figure~\ref{fig:bayesian_agent_architecture} summarizes the pipeline: a compatible harness emits verified trajectories, the Bayesian layer updates a skill registry and selects an action, and the resulting skill context is returned to the backend. The audit state remains separate from the model-facing text.

\subsection{Problem Formulation}

Let \(M_\theta\) be a frozen LLM or agentic model and let \(C_t\) denote the inference environment supplied at task \(t\). We decompose it as
\begin{equation}
  C_t=(P_t,R_t,A_t,V_t),
  \label{eq:environment_decomposition}
\end{equation}
where \(P_t\) is the prompt and skill text, \(R_t\) is retrieved or remembered context, \(A_t\) is the tool/action interface, and \(V_t\) is the verifier or feedback channel. Prompt, context, and harness engineering are therefore interventions on one conditional decision environment rather than unrelated heuristics.

Let \(h_k\) be a reusable skill or SOP, \(x_t\) a task, and \(y_t\in\{0,1\}\) the externally verified outcome. Let \(z_t^{-}\) denote the task and harness signature available before execution. The local reliability of the skill is
\begin{equation}
  p_{k,t}=P(y_t=1\mid M_\theta,C_t,h_k,z_t^{-}).
  \label{eq:skill_reliability}
\end{equation}
The model parameters stay fixed; optimization acts on the external conditions under which the model is run. Given belief state \(B_t\), the harness chooses an intervention \(\delta_t\) from a restricted action set \(\Delta\):
\begin{equation}
\begin{aligned}
  S_\delta&=P(y_t=1\mid M_\theta,C_t^\delta,h_k,z_t^{-}),\\
  \delta_t^\star&=\arg\max_{\delta\in\Delta}
  \mathbb{E}_{B_t}\left[S_\delta-
  \lambda_{\mathrm{cost}}\operatorname{Cost}(C_t^\delta)\right].
\end{aligned}
\label{eq:environment_decision}
\end{equation}
The implementation uses a conservative policy over skill text and guardrails, rather than searching the unrestricted space of prompts or harness programs.

\subsection{Finite-Sample Shrinkage}
\label{sec:bayes_freq_theory}

Consider one local skill cell \(\omega=(h_k,z)\), with unknown success probability \(\theta_\omega\). For \(n>0\) verified outcomes and \(S\) successes, the raw-frequency controller uses
\begin{equation}
  q_F=\frac{S}{n}.
  \label{eq:freq_local}
\end{equation}
Bayesian-Agent places \(\theta_\omega\sim\operatorname{Beta}(\alpha_0,\beta_0)\) and uses the posterior mean
\begin{equation}
  q_B=\mathbb{E}[\theta_\omega\mid S,n]
  =\frac{\alpha_0+S}{\alpha_0+\beta_0+n}.
  \label{eq:bayes_local}
\end{equation}
The assumptions are conditional exchangeability of verified Bernoulli outcomes within \(\omega\), positive prior pseudo-counts, and squared loss for the reliability estimate used by the skill policy. The prior is a regularizer for sparse evidence, not a guarantee of better performance.

\paragraph{Finite-sample result.}
Let \(A=\alpha_0+\beta_0\) and evaluate both estimators at a fixed but unknown \(\theta_\omega\). Since \(S\mid\theta_\omega\sim\operatorname{Binomial}(n,\theta_\omega)\), their fixed-parameter risks are
\begin{equation}
\begin{aligned}
  R_F(\theta_\omega)&=
  \mathbb{E}[(q_F-\theta_\omega)^2\mid\theta_\omega]\\
  &=\frac{\theta_\omega(1-\theta_\omega)}{n},\\
  R_B(\theta_\omega)&=
  \mathbb{E}[(q_B-\theta_\omega)^2\mid\theta_\omega]\\
  &=\frac{n\theta_\omega(1-\theta_\omega)+
  (\alpha_0-A\theta_\omega)^2}{(A+n)^2}.
\end{aligned}
\label{eq:fixed_theta_risk}
\end{equation}
Thus posterior shrinkage trades prior-directed bias for lower sampling variance. With the default \((\alpha_0,\beta_0)=(1,1)\),
\begin{equation}
\begin{aligned}
  R_B(\theta_\omega)&\leq R_F(\theta_\omega)\\
  &\Longleftrightarrow\quad
  \theta_\omega(1-\theta_\omega)\geq\frac{n}{8n+4}.
\end{aligned}
  \label{eq:uniform_prior_interval}
\end{equation}
The posterior mean therefore has lower fixed-parameter squared risk over an interior interval of reliabilities, while the empirical rate can be preferable near the boundaries when the prior is mismatched. Appendix~\ref{sec:appendix_bayes_freq_derivation} gives the derivation and the corresponding interval. This is the theoretical basis for using posterior belief as a cautious finite-sample state; the benchmark control tests its operational value.

\subsection{Trajectory Evidence and Execution Modes}

Bayesian-Agent updates beliefs only from externally verified trajectories. We represent a trajectory as
\begin{equation}
  e_t=(x_t,h_k,c_t,y_t,u_t,\tau_t,\ell_t,r_t,m_t),
\end{equation}
where \(c_t\) is task context, \(u_t\) total token cost, \(\tau_t\) turn count, \(\ell_t\) elapsed time, \(r_t\) a verifier-derived failure mode, and \(m_t\) short scalar metadata. After verification, the evidence feature map is
\begin{equation}
\begin{aligned}
  z_t^{+}=g(e_t)&=\big(c_t,r_t,b_u(u_t),b_\tau(\tau_t),\\
  &\qquad b_\ell(\ell_t),m_t^{\leq80}\big),
\end{aligned}
  \label{eq:evidence_features}
\end{equation}
where \(b_u,b_\tau,b_\ell\) are fixed buckets. The superscript \(+\) marks post-execution evidence: failure modes and runtime buckets guide later audits and repairs, while only components available in \(z_t^{-}\) are queried before the current run.

The schema supports two modes. In \textbf{full} mode, the registry starts empty and is updated after each task. In \textbf{incremental} mode, an existing verified run initializes the registry and only failed tasks are rerun with posterior-guided context. Full mode tests online evolution from scratch; incremental mode measures plug-in repair.

\subsection{Categorical Bayesian Evidence Model}

For skill \(h_k\), let \(D_{k,t}\) be its observed trajectories, \(N_{k,\ell}\) the number with outcome \(\ell\), and \(N_{k,j,\ell,v}\) the count for feature \(j\) taking value \(v\) under outcome \(\ell\). With Laplace smoothing \(\lambda=1\),
\begin{equation}
  \pi_{k,t}(\ell)=
  \frac{N_{k,\ell}+\lambda}
  {\sum_{\ell'\in\{0,1\}}N_{k,\ell'}+2\lambda}.
  \label{eq:class_prior}
\end{equation}
For a categorical value \(z_j=v\),
\begin{equation}
  \theta_{k,j,t}^{(\ell)}(v)=
  \frac{N_{k,j,\ell,v}+\lambda}
  {\sum_{v'\in\mathcal{V}_{k,j,t}}N_{k,j,\ell,v'}+
  \lambda|\mathcal{V}_{k,j,t}\cup\{v\}|}.
  \label{eq:feature_likelihood}
\end{equation}
The factorized score and normalized success posterior are
\begin{equation}
  \widetilde p_{k,t}(\ell\mid z)=
  \pi_{k,t}(\ell)\prod_{j=1}^{m}\theta_{k,j,t}^{(\ell)}(z_j),
  \label{eq:factorized_likelihood}
\end{equation}
\begin{equation}
  s_{k,t}(z)=
  \frac{\widetilde p_{k,t}(1\mid z)}
  {\widetilde p_{k,t}(0\mid z)+\widetilde p_{k,t}(1\mid z)}.
  \label{eq:success_posterior}
\end{equation}
Products are evaluated in log space. For audit compatibility, the registry also stores the conjugate summary
\begin{equation}
\begin{aligned}
  \alpha_{k,t}&=\alpha_0+\sum_{e_i\in D_{k,t}}\mathbf{1}[y_i=1],\\
  \beta_{k,t}&=\beta_0+\sum_{e_i\in D_{k,t}}\mathbf{1}[y_i=0].
\end{aligned}
  \label{eq:beta_summary}
\end{equation}
whose mean is \(\alpha_{k,t}/(\alpha_{k,t}+\beta_{k,t})\). The experiments use the categorical posterior for context selection; the Beta summary supplies a compact audit and failure-dominance signal.

\subsection{Posterior-Guided Skill Actions}

The posterior drives five inspectable actions. Let \(F_k(r)\) count failure mode \(r\), \(\mathcal{C}_k\) be observed contexts, and \(s_k(\varnothing)\) the unconditioned success posterior. The default ordered policy is
\begin{equation}
\pi(B_k)=
\begin{cases}
E,&|D_k|=0,\\
R,&\beta_k\geq4\ \land\ s_k(\varnothing)<0.45,\\
P,&\max_rF_k(r)\geq2,\\
S,&|\mathcal{C}_k|\geq3\ \land\ |D_k|\geq4,\\
C,&|D_k|\geq3\ \land\ s_k(\varnothing)\geq0.72,\\
E,&\text{otherwise,}
\end{cases}
\label{eq:rewrite_policy}
\end{equation}
where \(E,R,P,S,C\) denote explore, retire, patch, split, and compress. The thresholds are fixed implementation defaults; their meanings and alternatives are documented in Appendix~\ref{sec:appendix_method}. The model-facing context contains executable guardrails and repeated failure-mode patches, while posterior numbers remain in the audit state.

\begin{algorithm}[t]
\caption{Bayesian-Agent Evolution}
\label{alg:bayesian_agent}
\begin{algorithmic}[1]
\Require Frozen agent \(M_\theta\), task set \(\mathcal{T}\), mode \(m\in\{\textsc{Full},\textsc{Incremental}\}\), optional verified trace \(R^0\)
\Ensure Evolved registry \(\mathcal{B}\), task outputs, and before/after records
\State Initialize \(\mathcal{B}\gets\emptyset\)
\If{\(m=\textsc{Incremental}\)}
  \ForAll{verified trajectory \(e\in R^0\)}
    \State \(z\gets g(e)\); update \(B_{h(e)}\) with Eqs.~\ref{eq:class_prior}--\ref{eq:success_posterior}
  \EndFor
  \State \(\mathcal{T}'\gets\{x\in\mathcal{T}:x\text{ failed in }R^0\}\)
\Else
  \State \(\mathcal{T}'\gets\mathcal{T}\)
\EndIf
\ForAll{task \(x_t\in\mathcal{T}'\)}
  \State Select skill \(h_k\) and posterior state \(B_k\); compute \(a_t\gets\pi(B_k)\) from available context
  \State Render guardrails and failure patches as model-facing context \(q_t\)
  \State Save before state; execute \(M_\theta\) through the harness to obtain \(o_t\)
  \State Verify \(o_t\) to obtain \(y_t\), costs, and failure mode \(r_t\)
  \State Construct \(e_t\), extract \(z_t^{+}=g(e_t)\), update \(B_k\), and save the after state
\EndFor
\end{algorithmic}
\end{algorithm}

The native backend implements a small chat/tool loop, workspace-scoped tools, usage accounting, and trajectory persistence. External backends use the same adapter boundary: they execute tasks and expose verified trajectory fields, while Bayesian-Agent owns evidence ingestion, posterior updates, action selection, skill rendering, and evolution records. This separation makes the skill layer portable across harnesses without assuming identical execution loops.

\section{Experiments}
\label{sec:exp}

\subsection{Setup}

We evaluate four questions. First, does a posterior belief state improve over a raw empirical-rate controller under sparse evidence? Second, do full online evolution and incremental repair improve the GenericAgent baseline? Third, does the same skill layer transfer across a harness-complexity ladder? Fourth, how does model scale interact with skill evolution? \textbf{GA} is the GenericAgent baseline, \textbf{BA-Full} evolves an empty registry online, and \textbf{BA-Inc} ingests a completed baseline trace and reruns only failed tasks.

The backend ablation compares the native backend, MiniSWEAgent, and Claude Code using full-sample baseline, full, and incremental runs with both DeepSeek backbones. The model ablation fixes MiniSWEAgent and Bayesian full mode on RealFin-Bench, and compares Qwen3.5 with the DeepSeek models.

The benchmark suite covers three kinds of agent behavior. SOP-Bench tests multi-step procedural execution over industrial SOPs \citep{sopbench}. Lifelong AgentBench evaluates whether agents can handle sequential tasks with reusable cross-task experience \citep{lifelongagentbench}. RealFin-Bench evaluates financial reasoning when important premises may be implicit or missing \citep{realfin}. For benchmark definitions and task grading, we follow the GenericAgent evaluation setup \citep{genericagent2026}.

We report task accuracy, input/output/total tokens, and solved tasks per million tokens. Table~\ref{tab:bayes_vs_freq} isolates the Bayesian-vs.-raw-frequency question; Table~\ref{tab:main_results} gives the main benchmark comparison. Full and baseline rows use full-run tokens, whereas BA-Inc reports repair-only tokens and final post-repair accuracy. Complete accounting appears in Appendix~\ref{sec:appendix_tokens} and Appendix~\ref{sec:appendix_backend_full_sample}.

\subsection{Bayesian vs. Raw-Frequency Skill Evolution}
\label{sec:bayes_vs_freq_exp}

The Bayesian-vs.-raw-frequency control runs both methods from scratch on 40 RealFin tasks. Raw frequency replaces the posterior with an unsmoothed empirical success rate; all other task and execution conditions are held fixed.

\begin{table*}[!t]
\centering
\caption{\textbf{Bayesian vs. raw-frequency skill evolution on RealFin-Bench.} All rows are full 40-task runs. Efficiency is solved tasks per million total tokens; raw frequency uses an unsmoothed empirical success rate. Higher is better for accuracy, success, and efficiency, and lower is better for token columns.}
\resizebox{\linewidth}{!}{
\begin{tabular}{lllcccccc}
\toprule
\textbf{Backend} & \textbf{Model} & \textbf{Method} & \textbf{Accuracy}\upmark & \textbf{Success}\upmark & \textbf{Input}\downmark & \textbf{Output}\downmark & \textbf{Total}\downmark & \textbf{Efficiency}\upmark \\
\midrule
GA & flash & Bayesian & 52.5\% & 21/40 & \bestmetric{2.90M} & \secondmetric{244k} & \bestmetric{3.15M} & \secondmetric{6.67} \\
GA & flash & Raw freq. & 52.5\% & 21/40 & \secondmetric{3.06M} & 244k & \secondmetric{3.31M} & 6.35 \\
GA & pro & Bayesian & \secondmetric{65.0\%} & \secondmetric{26/40} & 3.38M & 323k & 3.70M & \bestmetric{7.02} \\
GA & pro & Raw freq. & 57.5\% & 23/40 & 3.46M & 309k & 3.77M & 6.10 \\
Native BA & flash & Bayesian & \bestmetric{70.0\%} & \bestmetric{28/40} & 10.43M & 463k & 10.89M & 2.57 \\
Native BA & flash & Raw freq. & 37.5\% & 15/40 & 3.79M & \bestmetric{236k} & 4.03M & 3.72 \\
Native BA & pro & Bayesian & \bestmetric{70.0\%} & \bestmetric{28/40} & 9.33M & 579k & 9.91M & 2.83 \\
Native BA & pro & Raw freq. & 45.0\% & 18/40 & 3.86M & 299k & 4.16M & 4.32 \\
\bottomrule
\end{tabular}
}
\label{tab:bayes_vs_freq}
\end{table*}

\begin{figure*}[t]
\centering
\includegraphics[width=\linewidth]{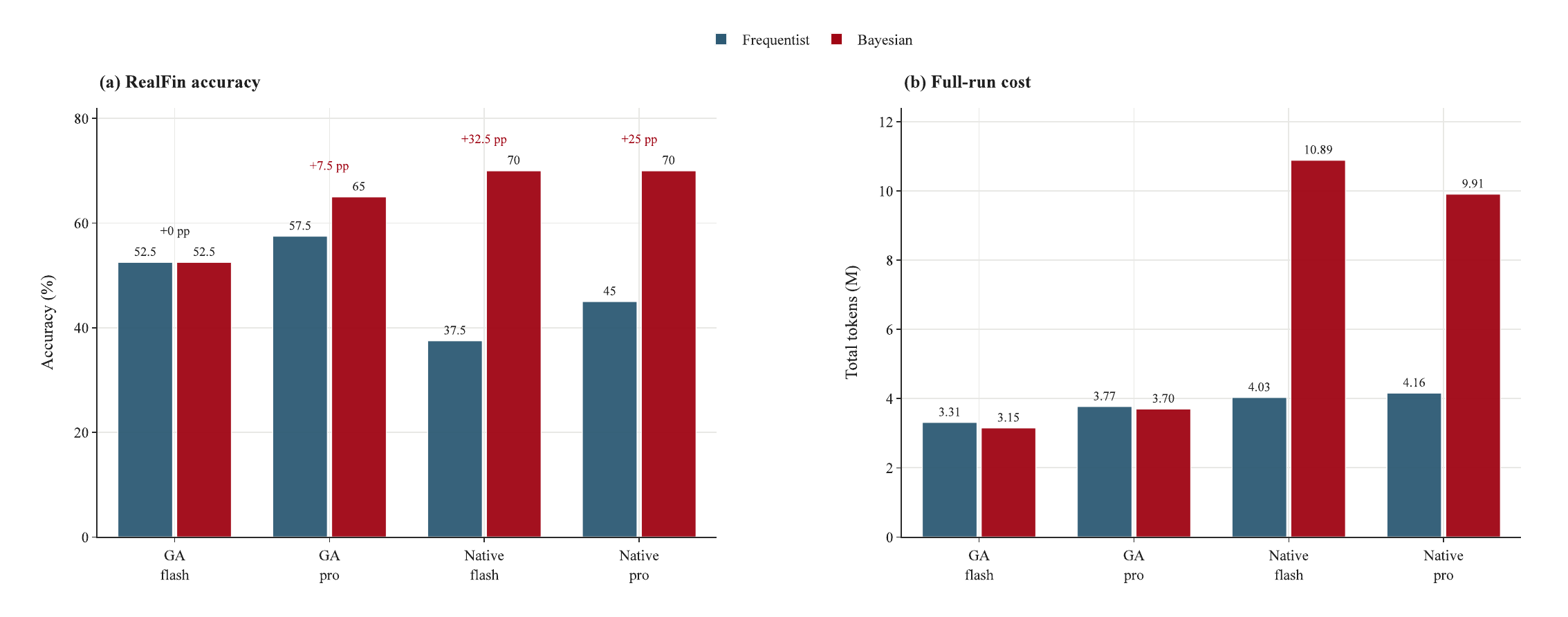}
\caption{\textbf{Bayesian-vs.-raw-frequency RealFin control.} Panel (a) compares completion accuracy and annotates Bayesian accuracy change over the raw-frequency controller; panel (b) compares full-run token cost. The native Bayesian rows spend more tokens, so the evidence supports stronger task completion under sparse evidence rather than lower cost.}
\label{fig:bayes_vs_freq}
\end{figure*}

Table~\ref{tab:bayes_vs_freq} and Figure~\ref{fig:bayes_vs_freq} show that posterior belief is most valuable in the native sparse-evidence setting. Bayesian-Agent reaches 28/40 tasks for both DeepSeek models, compared with 15/40 and 18/40 for raw frequency, while the GA-backed comparison is more mixed. The cheaper raw-frequency native runs often terminate before producing valid benchmark outputs, so token efficiency alone is misleading. Posterior smoothing helps when sparse evidence contains recoverable procedural failures, while raw rates remain competitive in other regimes.

\begin{table*}[!t]
\centering
\caption{\textbf{Main task completion and token efficiency.} GA and BA-Full use full-run tokens; BA-Inc uses repair-only tokens and reports final post-repair accuracy. Efficiency is solved tasks per million tokens for full runs and repaired successes per million repair tokens for BA-Inc. Best values are boldfaced within each dataset block.}
\scriptsize
\def\arraystretch{0.92}
\setlength{\tabcolsep}{0.42em}
\resizebox{1.0\linewidth}{!}{
\begin{tabular}{llccccc}
\toprule
\textbf{Agent} & \textbf{Model} & \textbf{Accuracy}\upmark & \textbf{Input Tokens}\downmark & \textbf{Output Tokens}\downmark & \textbf{Total Tokens}\downmark & \textbf{Efficiency}\upmark \\
\midrule
\rowcolor{gray!30}
\multicolumn{7}{c}{\textbf{SOP-Bench}} \\
GA & Claude Sonnet 4.6 & \bestmetric{100\%} & 2.02M & 53k & 2.08M & 0.48 \\
OpenClaw & Claude Sonnet 4.6 & \bestmetric{100\%} & 2.60M & 40k & 2.64M & 0.38 \\
Claude Code & Claude Sonnet 4.6 & 85\% & 1.23M & 23k & 1.25M & 0.68 \\
GA & deepseek-v4-flash & 80\% & 1.34M & 57k & 1.39M & 11.47 \\
BA-Full & deepseek-v4-flash & \secondmetric{95\%} & 1.20M & 60k & 1.26M & 15.13 \\
BA-Inc & deepseek-v4-flash & 95\% & \bestmetric{145k} & \bestmetric{8k} & \bestmetric{153k} & 19.63 \\
GA & deepseek-v4-pro & \bestmetric{100\%} & 857k & 45k & 902k & \bestmetric{22.18} \\
BA-Full & deepseek-v4-pro & \bestmetric{100\%} & 914k & 51k & 965k & 20.73 \\
BA-Inc & deepseek-v4-pro & \bestmetric{100\%} & 0 & 0 & 0 & -- \\
\midrule
\rowcolor{gray!30}
\multicolumn{7}{c}{\textbf{Lifelong AgentBench}} \\
GA & Claude Sonnet 4.6 & \bestmetric{100\%} & 222k & 20k & 241k & 4.15 \\
OpenClaw & Claude Sonnet 4.6 & 70\% & 1.43M & 21k & 1.45M & 0.48 \\
Claude Code & Claude Sonnet 4.6 & 75\% & 800k & 14k & 814k & 0.92 \\
GA & deepseek-v4-flash & 90\% & 649k & 42k & 690k & \secondmetric{26.07} \\
BA-Full & deepseek-v4-flash & 85\% & 634k & 40k & 674k & 25.23 \\
BA-Inc & deepseek-v4-flash & \bestmetric{100\%} & \bestmetric{78k} & \bestmetric{6k} & \bestmetric{84k} & 23.85 \\
GA & deepseek-v4-pro & \bestmetric{100\%} & 385k & 34k & 418k & \bestmetric{47.82} \\
BA-Full & deepseek-v4-pro & \bestmetric{100\%} & 405k & 35k & 439k & 45.53 \\
BA-Inc & deepseek-v4-pro & \bestmetric{100\%} & 0 & 0 & 0 & -- \\
\midrule
\rowcolor{gray!30}
\multicolumn{7}{c}{\textbf{RealFin-Bench}} \\
GA & Claude Sonnet 4.6 & 65\% & \bestmetric{102k} & 12k & \bestmetric{114k} & 5.70 \\
Claude Code & Claude Opus 4.6 & 60\% & 290k & 17k & 307k & 1.95 \\
Claude Code & Claude Sonnet 4.6 & 55\% & 226k & 12k & 238k & 2.31 \\
OpenClaw & Claude Sonnet 4.6 & 35\% & 249k & \bestmetric{2k} & 251k & 1.39 \\
Codex & GPT-5.4 & 60\% & 838k & 54k & 892k & 0.67 \\
GA & deepseek-v4-flash & 45\% & 3.98M & 262k & 4.24M & 4.24 \\
BA-Full & deepseek-v4-flash & 52\% & 2.90M & 244k & 3.15M & 6.67 \\
BA-Inc & deepseek-v4-flash & 65\% & 1.89M & 126k & 2.02M & 3.96 \\
GA & deepseek-v4-pro & 60\% & 3.40M & 317k & 3.72M & 6.46 \\
BA-Full & deepseek-v4-pro & 65\% & 3.38M & 323k & 3.70M & \bestmetric{7.02} \\
BA-Inc & deepseek-v4-pro & \bestmetric{68\%} & 1.59M & 130k & 1.72M & 1.74 \\
\bottomrule
\end{tabular}
}
\label{tab:main_results}
\end{table*}

\begin{figure*}[!b]
\centering
\includegraphics[width=0.96\linewidth]{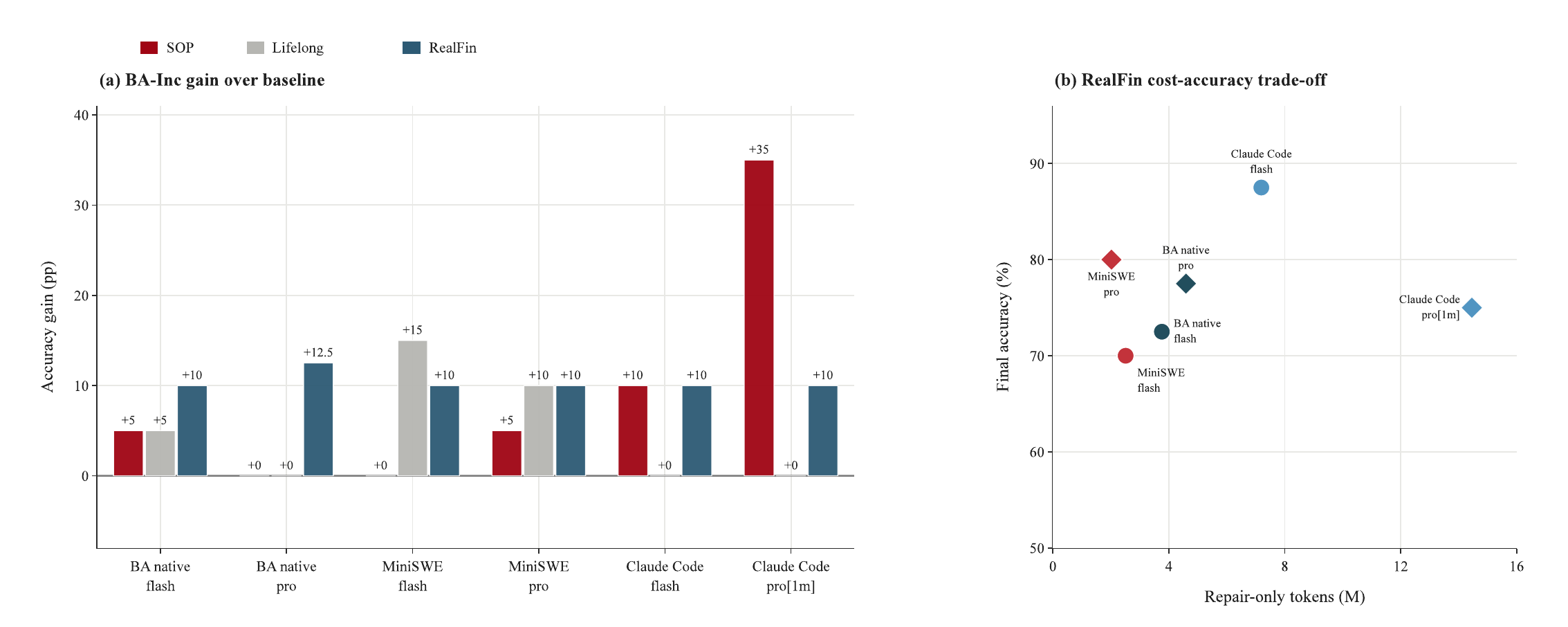}
\caption{\textbf{Backend ablation across the harness-complexity ladder.} Panel (a) shows BA-Inc final accuracy gain over the corresponding baseline for each backend, model, and benchmark. Panel (b) plots the RealFin repair-only cost versus final accuracy trade-off across backend/model choices.}
\label{fig:backend_ablation}
\end{figure*}

\subsection{Main Results}

Table~\ref{tab:main_results} shows that the largest gains occur when baseline failures are recoverable. With \texttt{flash}, BA-Full improves SOP-Bench from 80\% to 95\% and RealFin-Bench from 45\% to 52\%, while BA-Inc reaches 95\%, 100\%, and 65\% on SOP-Bench, Lifelong AgentBench, and RealFin-Bench. Online evolution is not monotonic: BA-Full falls from 90\% to 85\% on Lifelong AgentBench with \texttt{flash}, whereas incremental repair reaches 100\%. The pro setting is partly saturated on SOP-Bench and Lifelong AgentBench but still gains on RealFin-Bench, where BA-Inc reaches 68\%. Appendix~\ref{sec:appendix_visual_analysis} provides a visual summary of these benchmark-level trends (Figure~\ref{fig:bayesian_results}).

\subsection{Backend Ablation}
\label{sec:backend_ablation}

\begin{table*}[!b]
\centering
\caption{\textbf{Backend ablation across an increasing harness-complexity ladder.} Each cell reports baseline / BA-Full / BA-Inc accuracy. Detailed token accounting and efficiency are in Appendix~\ref{sec:appendix_backend_full_sample}. Best values are boldfaced within each benchmark column.}
\resizebox{\linewidth}{!}{
\begin{tabular}{llccccccccc}
\toprule
\textbf{Backend} & \textbf{Model} &
\multicolumn{3}{c}{\textbf{SOP-Bench}} &
\multicolumn{3}{c}{\textbf{Lifelong}} &
\multicolumn{3}{c}{\textbf{RealFin}} \\
\cmidrule(lr){3-5}\cmidrule(lr){6-8}\cmidrule(lr){9-11}
 & & Base\upmark & Full\upmark & Inc\upmark & Base\upmark & Full\upmark & Inc\upmark & Base\upmark & Full\upmark & Inc\upmark \\
\midrule
BA native & flash & 95\% & \bestmetric{100\%} & \bestmetric{100\%} & 95\% & \bestmetric{100\%} & \bestmetric{100\%} & 62.5\% & 70\% & 72.5\% \\
BA native & pro & \bestmetric{100\%} & \bestmetric{100\%} & \bestmetric{100\%} & \bestmetric{100\%} & \bestmetric{100\%} & \bestmetric{100\%} & 65\% & 70\% & 77.5\% \\
MiniSWEAgent & flash & \bestmetric{100\%} & 95\% & \bestmetric{100\%} & 85\% & 95\% & \bestmetric{100\%} & 60\% & 55\% & 70\% \\
MiniSWEAgent & pro & 95\% & \bestmetric{100\%} & \bestmetric{100\%} & 90\% & \bestmetric{100\%} & \bestmetric{100\%} & 70\% & 70\% & 80\% \\
Claude Code & flash & 90\% & \bestmetric{100\%} & \bestmetric{100\%} & \bestmetric{100\%} & \bestmetric{100\%} & \bestmetric{100\%} & \bestmetric{77.5\%} & \bestmetric{80\%} & \bestmetric{87.5\%} \\
Claude Code & pro[1m] & 65\% & 95\% & \bestmetric{100\%} & \bestmetric{100\%} & \bestmetric{100\%} & \bestmetric{100\%} & 65\% & 67.5\% & 75\% \\
\bottomrule
\end{tabular}
}
\label{tab:backend_ablation}
\end{table*}

Table~\ref{tab:backend_ablation} and Figure~\ref{fig:backend_ablation} show a consistent incremental-repair advantage across the harness-complexity ladder. On RealFin-Bench, BA-Inc improves the native backend to 72.5\%/77.5\%, MiniSWEAgent to 70\%/80\%, and Claude Code to 87.5\%/75\% for \texttt{flash}/\texttt{pro}. Full evolution is less stable in a few settings, including MiniSWEAgent \texttt{flash}. Claude Code + \texttt{flash} is strongest in this full-sample ablation, but its 7.19M repair tokens and 49.41M-token baseline show that the result is a backend/model/contract match, not evidence that \texttt{flash} is intrinsically better than \texttt{pro}. MiniSWEAgent + \texttt{pro} offers the most favorable RealFin repair cost among the high-performing incremental rows. Backend choice should therefore jointly consider completion, cost, contract adherence, and execution-stack fit.

\subsection{Model Scaling Ablation}
\label{sec:model_scaling}

The model-scaling ablation fixes MiniSWEAgent, Bayesian full mode, and 40 RealFin tasks while varying the model; the complete table and failure-shape visualization are deferred to Appendix~\ref{sec:appendix_model_scaling} (Table~\ref{tab:model_scaling} and Figure~\ref{fig:model_scaling}). The results show a conditional scaling signal: accuracy rises from 45.0\% for Qwen3.5-35B-A3B to 55.0\% for DeepSeek V4 Flash and 70.0\% for DeepSeek V4 Pro. The Qwen3.5-122B-A10B exception is dominated by missing output files, indicating an adapter-contract failure rather than a clean test of reasoning scale.

\subsection{Discussion}
\label{sec:discussion}

The results support a narrow operational conclusion. Posterior-guided incremental repair is most useful when a harness exposes verified, reusable failure evidence and gives the skill layer a place to act. The backend and model ablations show that stronger execution can improve completion, but only when the harness contract turns behavior into valid artifacts; otherwise a larger model can produce less useful evidence.

This benefit has clear limits. Full online evolution can be order-sensitive under sparse evidence, and empirical rates may be adequate when observations are plentiful and stable. Bayesian-Agent is best suited to repeated tasks with verifiers, recurring procedural failures, and a controllable skill interface; it is less suitable for one-off subjective tasks, highly nonstationary environments, or failures caused by unavailable tools or data. Detailed costs, order sensitivity, and evolution records are deferred to the appendices \ref{sec:appendix_tokens}, \ref{sec:appendix_backend_full_sample}, \ref{sec:appendix_order_sensitivity}.

\section{Conclusion}
\label{sec:conclusion}

This paper presented Bayesian-Agent, a native and cross-harness framework that treats reusable agent skills and SOPs as posterior belief states. Verified trajectories update a compact evidence model, and a conservative policy turns that state into inspectable skill actions and executable failure-mode patches. The finite-sample analysis explains why shrinkage can stabilize sparse reliability estimates, while the Bayesian-vs.-raw-frequency control tests whether that stability improves task completion.
Our main contribution is Bayesian-Agent, a native and cross-harness skill-evolution framework that replaces stateless empirical patching with auditable posterior-guided updates over reusable agent skills under sparse verified evidence.

The results support a conditional practical lesson: incremental repair is most effective when repeated procedural failures are verifiable and the harness exposes a controllable skill interface. 
Future work includes richer Bayesian decision policies, broader adapters, and transfer of skill beliefs across models and deployments.

\clearpage
\bibliography{custom}

\clearpage
\appendix
\section*{Appendices}

This supplementary material is organized to make the main results and design choices traceable. Appendix~\ref{sec:appendix_method} and Appendix~\ref{sec:appendix_bayes_freq_derivation} expand the Bayesian modeling details behind the method. Appendix~\ref{sec:appendix_tokens}--\ref{sec:appendix_model_scaling} give the complete accounting behind the experimental tables, including the visual summaries deferred from the main paper, and Appendix~\ref{sec:appendix_order_sensitivity} tests robustness to within-benchmark task order on the native backend. Appendix~\ref{sec:bayes_freq_case} and Appendix~\ref{sec:case_study} then expose representative task-level and skill-text traces, so the reader can inspect how verified failures become posterior evidence and model-facing harness edits.

\begin{itemize}
    \item Appendix~\ref{sec:appendix_method}: implementation details for evidence features and policy boundaries
    \item Appendix~\ref{sec:appendix_bayes_freq_derivation}: the local finite-sample shrinkage and risk derivation
    \item Appendix~\ref{sec:appendix_tokens}: repair-only and cumulative token accounting
    \item Appendix~\ref{sec:appendix_backend_full_sample}: complete full-sample backend tables with input/output/total tokens
    \item Appendix~\ref{sec:appendix_visual_analysis}: visual summary of Bayesian-Agent benchmark results
    \item Appendix~\ref{sec:appendix_model_scaling}: RealFin model-scaling tables, visualization, protocol, and interpretation boundary
    \item Appendix~\ref{sec:appendix_order_sensitivity}: robustness to task execution order on the native backend
    \item Appendix~\ref{sec:bayes_freq_case}: task-level Bayesian-vs.-raw-frequency RealFin failure patterns
    \item Appendix~\ref{sec:case_study}: before/after skill-evolution traces and full model-facing skill texts
    \item Appendix~\ref{sec:appendix_admin}: limitations, ethics, and AI-use disclosure
\end{itemize}

\section{Additional Method Details}
\label{sec:appendix_method}

\paragraph{Evidence features.}
The deployed Bayesian evidence model uses compact categorical features rather than a large latent state. Each trajectory contributes benchmark context, failure mode, token, turn, and latency buckets, together with selected scalar metadata. Raw costs, verifier outputs, and task metadata are retained for audit. This design keeps posterior updates cheap while exposing recurring failure signatures.

\paragraph{Policy boundary.}
The policy in Table~\ref{tab:policy} uses fixed implementation thresholds. A downstream system may replace these thresholds with contextual bandits or a richer decision policy over the same evidence schema. The stable interface is that verified evidence updates an auditable posterior and emits explicit actions such as patch, split, compress, retire, or explore.

\begin{table*}[t]
\centering
\small
\setlength{\tabcolsep}{0.45em}
\begin{tabular}{p{0.24\linewidth}p{0.64\linewidth}}
\toprule
Symbol & Meaning \\
\midrule
\(\omega=(h_k,z)\) & Local skill cell: skill \(h_k\) under trajectory signature \(z\). \\
\(\theta_\omega\) & Unknown success probability of the frozen model under cell \(\omega\). \\
\(y_i\) & Verified binary task outcome from the grader or output contract. \\
\(n,S\) & Number of observations and successes in the local cell. \\
\(q_F,q_B\) & Empirical-rate estimate and Beta-shrunk posterior mean. \\
\(\alpha_0,\beta_0\) & Positive Beta-prior pseudo-counts. \\
\bottomrule
\end{tabular}
\caption{Notation for the finite-sample risk analysis.}
\label{tab:bayes_freq_notation}
\end{table*}

\begin{table*}[t]
\centering
\small
\begin{tabular}{p{0.19\linewidth}p{0.70\linewidth}}
\toprule
Action & Default trigger \\
\midrule
Explore & No observations are available, or the posterior remains uncertain. \\
Patch & The same failure mode appears at least twice. \\
Split & At least three contexts and four observations suggest a heterogeneous skill. \\
Compress & At least three observations and estimated success probability at least 0.72. \\
Retire & Failure evidence dominates: \(\beta\geq4\) and estimated success probability below 0.45. \\
\bottomrule
\end{tabular}
\caption{Default posterior-guided skill actions. Thresholds are implementation defaults.}
\label{tab:policy}
\end{table*}

\section{Finite-Sample Risk Derivation for Sparse Skill Cells}
\label{sec:appendix_bayes_freq_derivation}

This appendix expands the finite-sample derivation in Section~\ref{sec:bayes_freq_theory}. The analysis compares two concrete skill-maintenance states used in the paper: a raw empirical-rate estimate \(q_F=S/n\), and a Beta-shrunk posterior mean \(q_B=(\alpha_0+S)/(\alpha_0+\beta_0+n)\). This comparison matches the experimental control, which replaces Bayesian-Agent's posterior state with an unsmoothed empirical success-rate controller. Consider one fixed skill cell \(\omega=(h_k,z)\), where \(h_k\) is a reusable harness skill and \(z\) is a discrete trajectory feature signature. Let \(\theta_\omega\) denote the unknown probability that the frozen model succeeds when the harness uses this fixed skill version under this cell. We observe \(n\) verified Bernoulli outcomes \(y_1,\ldots,y_n\), with \(S=\sum_i y_i\) successes.

\paragraph{Model assumptions.}
The local derivation uses three assumptions. First, after fixing the model, backend, skill version, and feature cell, outcomes are conditionally exchangeable given \(\theta_\omega\):
\begin{equation}
  y_i\mid\theta_\omega \sim \mathrm{Bernoulli}(\theta_\omega).
\end{equation}
Second, the harness has a prior belief
\begin{equation}
  \theta_\omega \sim \mathrm{Beta}(\alpha_0,\beta_0),
\end{equation}
where \(\alpha_0,\beta_0>0\). Third, a controller reports a reliability estimate \(q\in[0,1]\) that is evaluated by squared prediction loss \(L(q,\theta_\omega)=(q-\theta_\omega)^2\). Squared loss gives a compact local criterion for comparing how raw empirical rates and posterior shrinkage behave under sparse evidence.

\paragraph{Posterior update.}
The likelihood of \(S\) successes in \(n\) observations is proportional to
\begin{equation}
  p(D\mid\theta_\omega) \propto
  \theta_\omega^{S}(1-\theta_\omega)^{n-S}.
\end{equation}
Multiplying by the Beta prior gives
\begin{equation}
\begin{aligned}
  p(\theta_\omega\mid D)
  &\propto
  \theta_\omega^{S+\alpha_0-1}
  (1-\theta_\omega)^{n-S+\beta_0-1}\\
  &=
  \mathrm{Beta}(\alpha_0+S,\beta_0+n-S).
\end{aligned}
\end{equation}
Therefore the Bayesian reliability estimate used by the derivation is the posterior mean
\begin{equation}
  q_B=
  \frac{\alpha_0+S}{\alpha_0+\beta_0+n}.
\end{equation}
The empirical-rate estimate is \(q_F=S/n\) when \(n>0\), and is undefined without a separate backoff rule when \(n=0\).

\paragraph{Fixed-\(\theta_\omega\) sampling risk.}
We also evaluate both estimators under the same fixed true reliability \(\theta_\omega\). Since \(S\mid\theta_\omega\sim\mathrm{Binomial}(n,\theta_\omega)\),
\begin{equation}
\begin{aligned}
  \mathbb{E}[q_F\mid\theta_\omega]=\theta_\omega,
  \\
  \mathrm{Var}(q_F\mid\theta_\omega)=
  \frac{\theta_\omega(1-\theta_\omega)}{n}.
\end{aligned}
\end{equation}
Thus the empirical-rate estimator has sampling risk
\begin{equation}
  R_F(\theta_\omega)
  =
  \mathbb{E}\left[(q_F-\theta_\omega)^2\mid\theta_\omega\right]
  =
  \frac{\theta_\omega(1-\theta_\omega)}{n}.
\end{equation}
Let \(A=\alpha_0+\beta_0\). For the posterior mean,
\begin{equation}
\begin{aligned}
  \mathbb{E}[q_B\mid\theta_\omega]=
  \frac{\alpha_0+n\theta_\omega}{A+n},
  \\
  \mathrm{Var}(q_B\mid\theta_\omega)=
  \frac{n\theta_\omega(1-\theta_\omega)}{(A+n)^2}.
\end{aligned}
\end{equation}
Its bias is
\begin{equation}
  \mathbb{E}[q_B\mid\theta_\omega]-\theta_\omega
  =
  \frac{\alpha_0-A\theta_\omega}{A+n}.
\end{equation}
Therefore
\begin{equation}
\begin{aligned}
  R_B(\theta_\omega)
  &=
  \mathbb{E}\left[(q_B-\theta_\omega)^2\mid\theta_\omega\right]\\
  &=
  \frac{
    n\theta_\omega(1-\theta_\omega)
    +(\alpha_0-A\theta_\omega)^2
  }{(A+n)^2}.
\end{aligned}
\end{equation}
The difference is a variance--bias tradeoff: \(q_B\) has lower sampling variance because its denominator is larger, but it is biased toward the prior mean \(\alpha_0/A\). With the default uniform prior \(\alpha_0=\beta_0=1\), algebra gives
\begin{equation}
  R_F(\theta_\omega)-R_B(\theta_\omega)
  =
  \frac{(8n+4)\theta_\omega(1-\theta_\omega)-n}
       {n(n+2)^2}.
\end{equation}
Thus
\begin{equation}
  \begin{gathered}
  R_B(\theta_\omega)\leq R_F(\theta_\omega)
  \\
  \Longleftrightarrow\quad
  \theta_\omega(1-\theta_\omega)\geq \frac{n}{8n+4}.
  \end{gathered}
\end{equation}
Equivalently,
\begin{equation}
\begin{aligned}
  \theta_\omega&\in[\ell_n,u_n],\\
  \ell_n&=\frac{1-\sqrt{(n+1)/(2n+1)}}{2},\\
  u_n&=\frac{1+\sqrt{(n+1)/(2n+1)}}{2}.
\end{aligned}
\end{equation}
For \(n=1\), this interval is approximately \([0.092,0.908]\); as \(n\) grows, it approaches \([0.146,0.854]\). The shrinkage estimator therefore has lower fixed-\(\theta_\omega\) mean-squared error for a broad interior range of skill reliabilities. The empirical rate has the advantage in edge cases where the true reliability is extremely close to \(0\) or \(1\) and the prior is poorly matched. This tradeoff is central for sparse skill cells: prior shrinkage prevents one observation from forcing an extreme decision, while the bias term records the cost of a mismatched prior.

\paragraph{Bayesian belief-state identity.}
After the evidence \(D\) is fixed, the same Beta-Bernoulli model gives the reliability estimate used by the harness under posterior squared loss. For any candidate estimate \(q\),
\begin{equation}
\begin{aligned}
  \rho(q\mid D)
  &=
  \mathbb{E}\left[(q-\theta_\omega)^2\mid D\right]\\
  &=
  \mathrm{Var}(\theta_\omega\mid D)
  +
  \left(q-\mathbb{E}[\theta_\omega\mid D]\right)^2.
\end{aligned}
\end{equation}
The variance term is independent of \(q\), so \(\rho(q\mid D)\) is minimized by \(q=q_B\). Comparing this posterior-loss value against the empirical estimate yields the standard Bayes-decision identity
\begin{equation}
\begin{aligned}
  &\rho(q_F\mid D)-\rho(q_B\mid D)\\
  &=
  (q_F-q_B)^2\\
  &=
  \left(
  \frac{\beta_0S-\alpha_0(n-S)}
       {n(\alpha_0+\beta_0+n)}
  \right)^2
  \geq 0.
\end{aligned}
\end{equation}
Within this posterior-loss criterion, the posterior mean is the coherent reliability estimate for the belief state used by the harness. The fixed-\(\theta_\omega\) sampling-risk calculation above then describes when this shrinkage also reduces mean-squared error under repeated sparse samples.

\paragraph{Connection to harness decisions.}
The deployed Bayesian-Agent embeds this scalar cell in a feature-conditioned policy with thresholds for patch, split, compress, and retire. The derivation explains the local decision principle behind that implementation. When evidence is scarce, the posterior mean provides controlled shrinkage and an explicit uncertainty state, while the empirical rate is a high-variance observation summary. Under sparse skill evidence with a reasonable prior, posterior belief is a more stable finite-sample decision state before the harness commits to durable skill edits. The direct control in Section~\ref{sec:bayes_vs_freq_exp} then tests whether this tradeoff improves task completion in a realistic agent benchmark.

\section{Additional Token Accounting}
\label{sec:appendix_tokens}

BA-Inc has two cost interpretations, and both are useful for different questions. The repair-only view measures the marginal cost of adding Bayesian-Agent after a completed baseline run; this is the quantity reported in the main task-completion table. The cumulative view adds the original baseline tokens and is the appropriate quantity when comparing end-to-end cost from scratch. In the GenericAgent main comparison on \texttt{deepseek-v4-flash}, the cumulative BA-Inc totals are 1.55M tokens for SOP-Bench, 774k for Lifelong AgentBench, and 6.26M for RealFin-Bench. On \texttt{deepseek-v4-pro}, the cumulative RealFin-Bench cost is 5.44M tokens. For pro SOP-Bench and Lifelong AgentBench, BA-Inc performs no repair because GA already solves all tasks. In the full-sample backend ablation, Tables~\ref{tab:backend_baseline_details}--\ref{tab:backend_incremental_details} report the full baseline, full Bayesian, and repair-only accounting used to separate marginal repair cost from total execution cost.

The GenericAgent flash repair passes use 153k, 84k, and 2.02M repair-only tokens on SOP-Bench, Lifelong AgentBench, and RealFin-Bench; the corresponding pro RealFin repair uses 1.72M. In the backend ablation, Claude Code + \texttt{flash} uses 366k, 0, and 7.19M repair tokens on the three benchmarks after baseline totals of 5.89M, 1.55M, and 49.41M. Claude Code + \texttt{pro[1m]} uses 977k, 0, and 14.45M repair tokens after baseline totals of 2.76M, 1.55M, and 27.03M. These figures explain why repair-only efficiency and cumulative end-to-end cost answer different deployment questions.

\section{Full-Sample Backend Experiment Details}
\label{sec:appendix_backend_full_sample}

Tables~\ref{tab:backend_baseline_details}--\ref{tab:backend_incremental_details} provide the complete full-sample backend ablation behind Section~\ref{sec:backend_ablation}. The rows follow the harness-complexity ladder used in the main text: BA native, MiniSWEAgent, and Claude Code. For baseline and full runs, efficiency is solved tasks per million total tokens. For incremental runs, efficiency is repaired successes per million repair-only tokens; ``--'' means that the baseline solved all tasks and no repair was required.

\begin{table*}[p]
\centering
\caption{\textbf{Full-sample backend baseline details.} Token columns follow the format input + output = total, with efficiency computed as solved tasks per million total tokens. All entries attaining a column best are boldfaced.}
\scriptsize
\def\arraystretch{0.92}
\setlength{\tabcolsep}{0.42em}
\resizebox{\linewidth}{!}{
\begin{tabular}{lllccccc}
\toprule
\textbf{Backend} & \textbf{Model} & \textbf{Benchmark} & \textbf{Accuracy}\upmark & \textbf{Input}\downmark & \textbf{Output}\downmark & \textbf{Total}\downmark & \textbf{Efficiency}\upmark \\
\midrule
\rowcolor{gray!30}
\multicolumn{8}{l}{\textbf{BA native}} \\
 & flash & SOP-Bench & 95\% & 1.00M & 45k & 1.05M & 18.10 \\
 & flash & Lifelong AgentBench & 95\% & 495k & 42k & 538k & 35.32 \\
 & flash & RealFin-Bench & 62.5\% & 9.84M & 454k & 10.29M & 2.43 \\
 & pro & SOP-Bench & \bestmetric{100\%} & 696k & 47k & 744k & 26.88 \\
 & pro & Lifelong AgentBench & \bestmetric{100\%} & 383k & 38k & 422k & 47.39 \\
 & pro & RealFin-Bench & 65\% & 8.99M & 549k & 9.54M & 2.73 \\
\rowcolor{gray!30}
\multicolumn{8}{l}{\textbf{MiniSWEAgent}} \\
 & flash & SOP-Bench & \bestmetric{100\%} & 961k & 56k & 1.02M & 19.61 \\
 & flash & Lifelong AgentBench & 85\% & 527k & 53k & 581k & 29.26 \\
 & flash & RealFin-Bench & 60\% & 5.36M & 370k & 5.73M & 4.19 \\
 & pro & SOP-Bench & 95\% & 861k & 47k & 907k & 20.95 \\
 & pro & Lifelong AgentBench & 90\% & \bestmetric{309k} & 40k & \bestmetric{349k} & \bestmetric{51.58} \\
 & pro & RealFin-Bench & 70\% & 5.63M & 473k & 6.10M & 4.59 \\
\rowcolor{gray!30}
\multicolumn{8}{l}{\textbf{Claude Code}} \\
 & flash & SOP-Bench & 90\% & 5.81M & 78k & 5.89M & 3.06 \\
 & flash & Lifelong AgentBench & \bestmetric{100\%} & 1.53M & 25k & 1.55M & 12.90 \\
 & flash & RealFin-Bench & 77.5\% & 48.57M & \bestmetric{14k} & 49.41M & 0.63 \\
 & \texttt{pro[1m]} & SOP-Bench & 65\% & 2.71M & 50k & 2.76M & 4.71 \\
 & \texttt{pro[1m]} & Lifelong AgentBench & \bestmetric{100\%} & 1.53M & 24k & 1.55M & 12.90 \\
 & \texttt{pro[1m]} & RealFin-Bench & 65\% & 26.53M & 509k & 27.03M & 0.96 \\
\bottomrule
\end{tabular}
}
\label{tab:backend_baseline_details}
\end{table*}

\begin{table*}[p]
\centering
\caption{\textbf{Full-sample backend Bayesian Full details.} Token columns follow the format input + output = total, with efficiency computed as solved tasks per million total tokens. All entries attaining a column best are boldfaced.}
\scriptsize
\def\arraystretch{0.92}
\setlength{\tabcolsep}{0.42em}
\resizebox{\linewidth}{!}{
\begin{tabular}{lllccccc}
\toprule
\textbf{Backend} & \textbf{Model} & \textbf{Benchmark} & \textbf{Accuracy}\upmark & \textbf{Input}\downmark & \textbf{Output}\downmark & \textbf{Total}\downmark & \textbf{Efficiency}\upmark \\
\midrule
\rowcolor{gray!30}
\multicolumn{8}{l}{\textbf{BA native}} \\
 & flash & SOP-Bench & \bestmetric{100\%} & 828k & 43k & 870k & 22.99 \\
 & flash & Lifelong AgentBench & \bestmetric{100\%} & 473k & 41k & 514k & 38.91 \\
 & flash & RealFin-Bench & 70\% & 10.43M & 463k & 10.89M & 2.57 \\
 & pro & SOP-Bench & \bestmetric{100\%} & 695k & 44k & 739k & 27.06 \\
 & pro & Lifelong AgentBench & \bestmetric{100\%} & 397k & 39k & 437k & 45.77 \\
 & pro & RealFin-Bench & 70\% & 9.33M & 579k & 9.91M & 2.83 \\
\rowcolor{gray!30}
\multicolumn{8}{l}{\textbf{MiniSWEAgent}} \\
 & flash & SOP-Bench & 95\% & 957k & 55k & 1.01M & 18.81 \\
 & flash & Lifelong AgentBench & 95\% & 452k & 46k & 498k & 38.15 \\
 & flash & RealFin-Bench & 55\% & 6.50M & 453k & 6.96M & 3.16 \\
 & pro & SOP-Bench & \bestmetric{100\%} & 905k & 48k & 953k & 20.99 \\
 & pro & Lifelong AgentBench & \bestmetric{100\%} & \bestmetric{294k} & 36k & \bestmetric{330k} & \bestmetric{60.61} \\
 & pro & RealFin-Bench & 70\% & 6.28M & 459k & 6.73M & 4.16 \\
\rowcolor{gray!30}
\multicolumn{8}{l}{\textbf{Claude Code}} \\
 & flash & SOP-Bench & \bestmetric{100\%} & 3.39M & 55k & 3.44M & 5.81 \\
 & flash & Lifelong AgentBench & \bestmetric{100\%} & 1.54M & \bestmetric{24k} & 1.57M & 12.74 \\
 & flash & RealFin-Bench & 80\% & 47.18M & 774k & 47.95M & 0.67 \\
 & \texttt{pro[1m]} & SOP-Bench & 95\% & 2.78M & 45k & 2.82M & 6.74 \\
 & \texttt{pro[1m]} & Lifelong AgentBench & \bestmetric{100\%} & 1.55M & \bestmetric{24k} & 1.57M & 12.74 \\
 & \texttt{pro[1m]} & RealFin-Bench & 67.5\% & 32.04M & 691k & 32.73M & 0.82 \\
\bottomrule
\end{tabular}
}
\label{tab:backend_full_details}
\end{table*}

\begin{table*}[p]
\centering
\caption{\textbf{Full-sample backend Bayesian Incremental details.} Token columns report repair-only input + output = total. Efficiency is repaired successes per million repair-only tokens; ``--'' indicates no failed baseline cases were rerun. All entries attaining a column best are boldfaced; zero-token inactive rows are excluded from cost highlights.}
\scriptsize
\def\arraystretch{0.92}
\setlength{\tabcolsep}{0.42em}
\resizebox{\linewidth}{!}{
\begin{tabular}{lllccccc}
\toprule
\textbf{Backend} & \textbf{Model} & \textbf{Benchmark} & \textbf{Final Accuracy}\upmark & \textbf{Input}\downmark & \textbf{Output}\downmark & \textbf{Total}\downmark & \textbf{Efficiency}\upmark \\
\midrule
\rowcolor{gray!30}
\multicolumn{8}{l}{\textbf{BA native}} \\
 & flash & SOP-Bench & \bestmetric{100\%} & 43k & \bestmetric{2k} & 45k & 22.22 \\
 & flash & Lifelong AgentBench & \bestmetric{100\%} & 60k & 5k & 65k & 15.38 \\
 & flash & RealFin-Bench & 72.5\% & 3.57M & 191k & 3.76M & 1.06 \\
 & pro & SOP-Bench & \bestmetric{100\%} & 0 & 0 & 0 & -- \\
 & pro & Lifelong AgentBench & \bestmetric{100\%} & 0 & 0 & 0 & -- \\
 & pro & RealFin-Bench & 77.5\% & 4.34M & 246k & 4.59M & 1.09 \\
\rowcolor{gray!30}
\multicolumn{8}{l}{\textbf{MiniSWEAgent}} \\
 & flash & SOP-Bench & \bestmetric{100\%} & 0 & \bestmetric{2k} & 0 & -- \\
 & flash & Lifelong AgentBench & \bestmetric{100\%} & 34k & \bestmetric{2k} & 36k & \bestmetric{83.33} \\
 & flash & RealFin-Bench & 70\% & 2.37M & 138k & 2.51M & 1.59 \\
 & pro & SOP-Bench & \bestmetric{100\%} & 46k & \bestmetric{2k} & 48k & 20.83 \\
 & pro & Lifelong AgentBench & \bestmetric{100\%} & \bestmetric{25k} & \bestmetric{2k} & \bestmetric{27k} & 74.07 \\
 & pro & RealFin-Bench & 80\% & 1.89M & 128k & 2.02M & 1.98 \\
\rowcolor{gray!30}
\multicolumn{8}{l}{\textbf{Claude Code}} \\
 & flash & SOP-Bench & \bestmetric{100\%} & 360k & 6k & 366k & 5.46 \\
 & flash & Lifelong AgentBench & \bestmetric{100\%} & 0 & 0 & 0 & -- \\
 & flash & RealFin-Bench & \secondmetric{87.5\%} & 7.04M & 152k & 7.19M & 0.56 \\
 & \texttt{pro[1m]} & SOP-Bench & \bestmetric{100\%} & 958k & 20k & 977k & 7.16 \\
 & \texttt{pro[1m]} & Lifelong AgentBench & \bestmetric{100\%} & 0 & 0 & 0 & -- \\
 & \texttt{pro[1m]} & RealFin-Bench & 75\% & 14.15M & 296k & 14.45M & 0.28 \\
\bottomrule
\end{tabular}
}
\label{tab:backend_incremental_details}
\end{table*}

\clearpage

\section{Visual Analysis of Bayesian-Agent Results}
\label{sec:appendix_visual_analysis}

This appendix provides the visual summary of the main DeepSeek-backbone comparison referenced in Section~\ref{sec:exp}. Panel (a) shows accuracy across GA, BA-Full, and BA-Inc, while panel (b) summarizes the final gain from incremental repair over the corresponding baseline. These plots complement the aggregate values in Table~\ref{tab:main_results}.

\begin{figure*}[t]
\centering
\includegraphics[width=\linewidth]{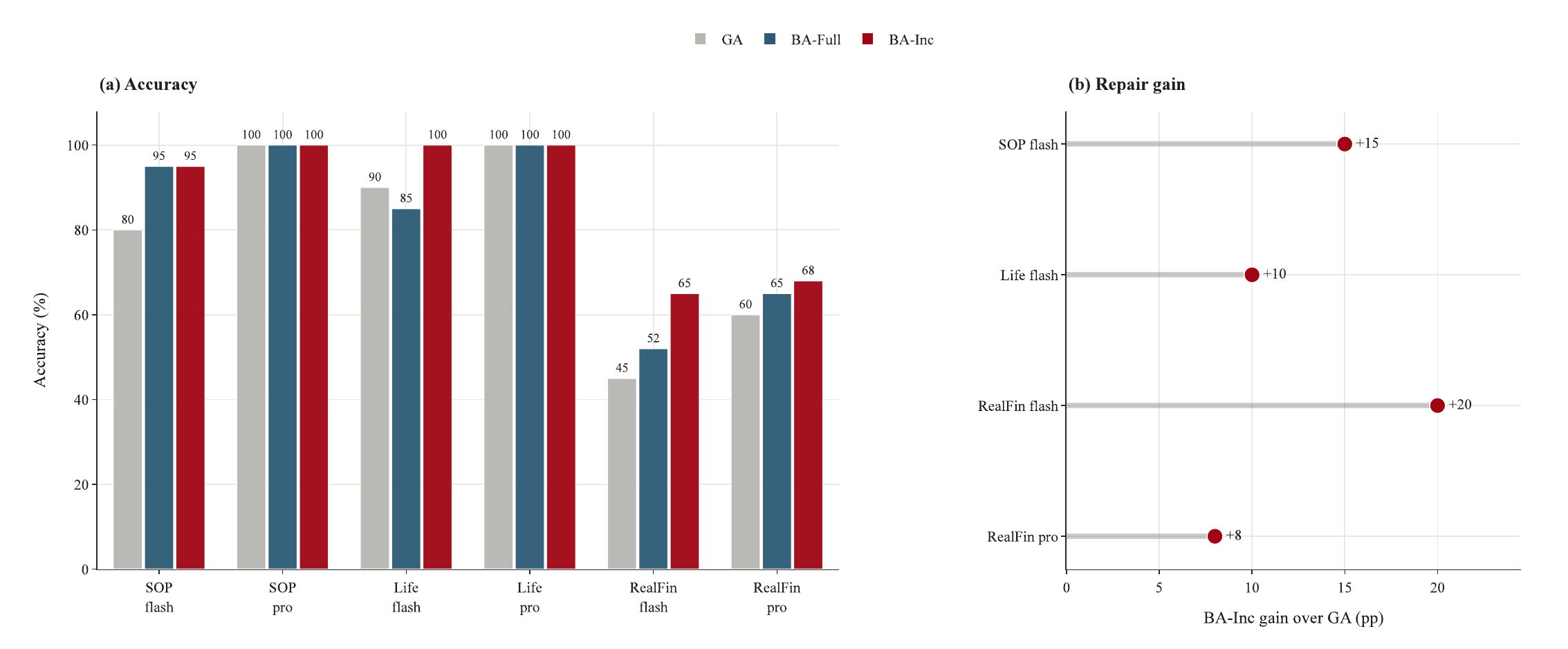}
\caption{\textbf{Visual analysis of Bayesian-Agent on DeepSeek backbones.} Panel (a) gives an annotated accuracy matrix for GA, BA-Full, and BA-Inc across benchmark-model settings. Panel (b) summarizes BA-Inc's final accuracy gain over GA for the non-zero repair settings, with parentheses showing repaired failures over baseline failures.}
\label{fig:bayesian_results}
\end{figure*}

\section{Model-Scaling RealFin Experiment Details}
\label{sec:appendix_model_scaling}

The model-scaling ablation in Section~\ref{sec:model_scaling} fixes the task, backend, and optimization mode so that the comparison focuses on model-and-adapter behavior. All rows use the \texttt{mini-swe-agent} backend, \texttt{bayesian\_full} evolution, RealFin-Bench with 40 tasks, \texttt{--max-turns 16}, and \texttt{--mini-swe-env-timeout 180}. The Qwen and DeepSeek rows use the same task, verifier, and Bayesian full protocol; only the model/provider configuration differs.

\begin{table*}[t]
\centering
\caption{\textbf{Model-scaling ablation on RealFin-Bench.} All rows use MiniSWEAgent with Bayesian full skill evolution. Efficiency is solved tasks per million total tokens. All entries attaining a best value for success, accuracy, token use, or efficiency are boldfaced; failure-shape counts are descriptive.}
\scriptsize
\def\arraystretch{0.94}
\setlength{\tabcolsep}{0.48em}
\resizebox{\linewidth}{!}{
\begin{tabular}{llcccccccc}
\toprule
\textbf{Model} & \textbf{Provider} & \textbf{Size} & \textbf{Success}\upmark & \textbf{Accuracy}\upmark & \textbf{Input}\downmark & \textbf{Output}\downmark & \textbf{Total}\downmark & \textbf{Efficiency}\upmark & \textbf{No File / Invalid}\downmark \\
\midrule
\texttt{qwen3.5-35b-a3b} & DashScope & 35B/A3B & 18/40 & 45.0\% & 6.95M & 370k & 7.32M & 2.46 & 10 / 12 \\
\texttt{qwen3.5-122b-a10b} & DashScope & 122B/A10B & 3/40 & 7.5\% & \bestmetric{3.19M} & \bestmetric{144k} & \bestmetric{3.34M} & 0.90 & 37 / 0 \\
\texttt{deepseek-v4-flash} & DeepSeek & 284B & 22/40 & 55.0\% & 6.50M & 453k & 6.96M & 3.16 & 9 / 9 \\
\texttt{deepseek-v4-pro} & DeepSeek & 1.6T & \bestmetric{28/40} & \bestmetric{70.0\%} & 6.28M & 459k & 6.73M & \bestmetric{4.16} & 7 / 5 \\
\bottomrule
\end{tabular}
}
\label{tab:model_scaling}
\end{table*}

\begin{figure*}[t]
\centering
\includegraphics[width=\linewidth]{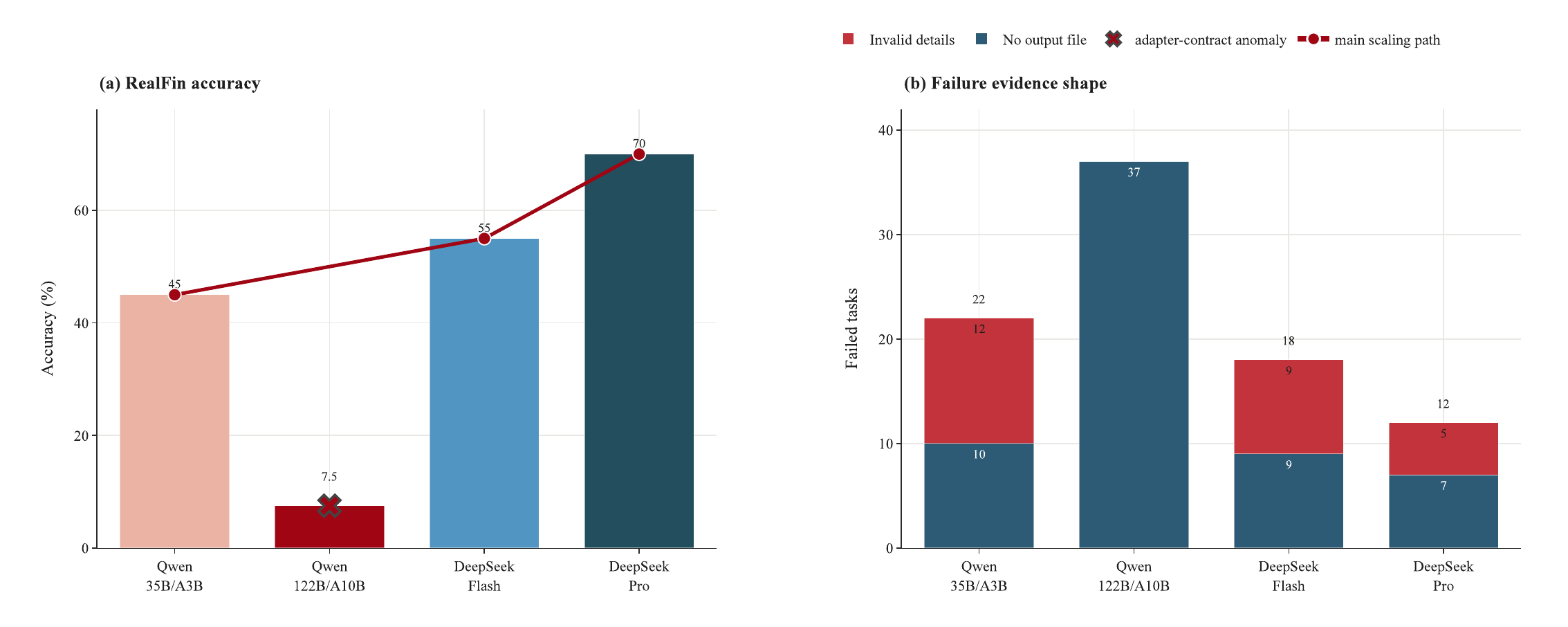}
\caption{\textbf{Model scaling on RealFin-Bench under MiniSWEAgent.} Panel (a) plots task accuracy against nominal size rank on a log scale. The connected path marks the main scaling signal from Qwen 35B-A3B to DeepSeek V4 Flash to DeepSeek V4 Pro, while Qwen 122B-A10B is shown as an adapter-contract anomaly. Panel (b) decomposes failed tasks by whether the requested output file was missing or a file was created but verifier details failed.}
\label{fig:model_scaling}
\end{figure*}

\paragraph{Interpretation boundary.}
This ablation focuses on model-and-adapter scaling, with parameter count only one factor. The cleanest positive signal is the path from \texttt{qwen3.5-35b-a3b} to \texttt{deepseek-v4-flash} to \texttt{deepseek-v4-pro}, where RealFin accuracy increases from 45.0\% to 55.0\% and 70.0\%. The \texttt{qwen3.5-122b-a10b} row isolates an adapter-contract bottleneck: all 37 failures have missing output files, pointing to artifact production and verification alignment as the dominant opportunity for improvement. The ablation therefore measures model-and-adapter behavior under a fixed MiniSWEAgent protocol, supporting a conditional scaling claim: larger or stronger models help Bayesian skill evolution when the harness contract turns their behavior into verifiable task artifacts.

\section{Robustness to Task Execution Order}
\label{sec:appendix_order_sensitivity}

We test whether the Bayesian-vs.-frequentist comparison is sensitive to task order. For each benchmark, we rerun the same task set in a different order while fixing the Bayesian-Agent native backend, \texttt{deepseek-v4-flash}, verifier, and full-evolution protocol. Each benchmark starts with a fresh skill state, so task order is the only changed factor.

\begin{table*}[t]
\centering
\caption{\textbf{Robustness to task execution order on the Bayesian-Agent native backend.} All runs use \texttt{deepseek-v4-flash}. Cells report solved tasks and accuracy under the original and reordered task sequences. \(\lvert\Delta\rvert\) is the absolute accuracy change caused by reordering, and \(\Delta_{\mathrm{BA-F}}\) is Bayesian-Agent's accuracy advantage over the frequentist raw-rate control under the reordered sequence.}
\small
\def\arraystretch{1.06}
\setlength{\tabcolsep}{0.85em}
\begin{tabular}{llcccc}
\toprule
\textbf{Benchmark} & \textbf{Method} & \textbf{Original Order}\upmark & \textbf{Reordered}\upmark & \textbf{\(\lvert\Delta\rvert\)}\downmark & \textbf{\(\Delta_{\mathrm{BA-F}}\)}\upmark \\
\midrule
SOP-Bench & Frequentist & 19/20 (95.0\%) & 19/20 (95.0\%) & \bestmetric{0.0} & -- \\
SOP-Bench & Bayesian-Agent & \bestmetric{20/20 (100\%)} & \bestmetric{20/20 (100\%)} & \bestmetric{0.0} & +5.0 \\
\cmidrule(lr){1-6}
Lifelong AgentBench & Frequentist & 19/20 (95.0\%) & 19/20 (95.0\%) & \bestmetric{0.0} & -- \\
Lifelong AgentBench & Bayesian-Agent & \bestmetric{20/20 (100\%)} & \bestmetric{20/20 (100\%)} & \bestmetric{0.0} & +5.0 \\
\cmidrule(lr){1-6}
RealFin-Bench & Frequentist & 15/40 (37.5\%) & 15/40 (37.5\%) & \bestmetric{0.0} & -- \\
RealFin-Bench & Bayesian-Agent & 28/40 (70.0\%) & 27/40 (67.5\%) & 2.5 & \bestmetric{+30.0} \\
\midrule
Overall & Frequentist & 53/80 (66.3\%) & 53/80 (66.3\%) & \bestmetric{0.0} & -- \\
Overall & Bayesian-Agent & 68/80 (85.0\%) & 67/80 (83.8\%) & 1.2 & +17.5 \\
\bottomrule
\end{tabular}
\label{tab:order_sensitivity}
\end{table*}

The comparison is stable under reordering. The frequentist control retains its original completion rate on all three benchmarks. Bayesian-Agent remains at 100\% on SOP-Bench and Lifelong AgentBench and changes by one task on RealFin-Bench, from 28/40 to 27/40. Across 80 tasks, its accuracy changes from 85.0\% to 83.8\%, while its reordered advantage remains 17.5 percentage points.

The small fluctuation does not arise from order sensitivity in the posterior arithmetic. For a fixed multiset of binary outcomes, both the empirical estimate \(S/n\) and the Beta posterior \(\operatorname{Beta}(\alpha_0+S,\beta_0+n-S)\) are permutation-invariant. However, skill evolution is closed loop: early observations can trigger skill actions that change the context and outcomes of later tasks. Thus, task order can change the evidence-generating path even when the update rule itself is invariant. The reported reorderings therefore characterize how the closed-loop evolution responds to alternative task sequences.

\section{Bayesian-vs.-Raw-Frequency RealFin Case Study}
\label{sec:bayes_freq_case}

\begin{figure*}[t]
\centering
\small
\begin{tcolorbox}[colback=gray!5!white, colframe=red!75!black,
title=RealFin Case Study: Bayesian Belief vs. Raw-Frequency Control, boxrule=0.3mm, width=\textwidth, arc=3mm, auto outer arc=true]
\centering
\begin{tikzpicture}[
  node distance=0.30cm and 0.30cm,
  bench/.style={font=\bfseries\scriptsize, align=right, text width=0.12\textwidth},
  box/.style={draw, rounded corners=1.2mm, align=left, text width=0.245\textwidth, inner sep=3pt, font=\scriptsize},
  freq/.style={box, fill=gray!9, draw=gray!65!black},
  bayes/.style={box, fill=green!6, draw=green!45!black},
  take/.style={box, fill=blue!5, draw=blue!55!black},
  arrow/.style={-{Latex[length=1.8mm]}, line width=0.35pt}
]
\node[bench] (flashlabel) {Native\\\texttt{flash}\\40 tasks};
\node[freq, right=of flashlabel] (flashfreq) {\textbf{Raw frequency}\\15/40 solved\\Empirical rate only; no prior smoothing.\\Early failures: blank OHLCV rows \(\times 5\), missing output files \(\times 6\).};
\node[bayes, right=of flashfreq] (flashbayes) {\textbf{Bayesian}\\28/40 solved\\Posterior belief keeps RealFin guardrails for cache reads, blank-row filtering, file creation, and format checks.};
\node[take, right=of flashbayes] (flashtake) {\textbf{Observation}\\+32.5 accuracy points.\\13 Bayesian-only successes; 0 raw-frequency-only successes.};
\draw[arrow] (flashfreq) -- (flashbayes);
\draw[arrow] (flashbayes) -- (flashtake);

\node[bench, below=1.05cm of flashlabel] (prolabel) {Native\\\texttt{pro}\\40 tasks};
\node[freq, right=of prolabel] (profreq) {\textbf{Raw frequency}\\18/40 solved\\Cheaper run, but many failures are missing-output, blank-field, or format non-answers.};
\node[bayes, right=of profreq] (probayes) {\textbf{Bayesian}\\28/40 solved\\Higher token budget becomes completed artifacts on complex RealFin indicator tasks.};
\node[take, right=of probayes] (protake) {\textbf{Observation}\\+25.0 accuracy points.\\11 Bayesian-only successes; 1 raw-frequency-only success.};
\draw[arrow] (profreq) -- (probayes);
\draw[arrow] (probayes) -- (protake);
\end{tikzpicture}
\end{tcolorbox}
\caption{Task-level case study for the native RealFin Bayesian-vs.-raw-frequency comparison. The raw-frequency controller is cheaper but often stops with missing output files or blank-field crashes. Bayesian-Agent spends more inference budget while retaining posterior-weighted execution guardrails, yielding higher completion on these sparse, expensive tasks.}
\label{fig:bayes_freq_case}
\end{figure*}
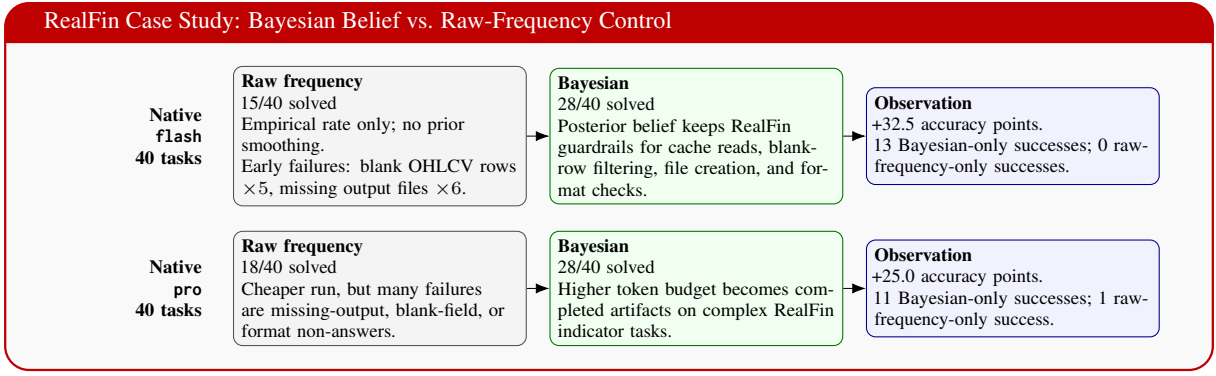

The native RealFin runs expose the largest Bayesian-vs.-raw-frequency gap because many failures occur before the benchmark can grade the intended financial computation. Figure~\ref{fig:bayes_freq_case} summarizes the task-level pattern. The raw-frequency controller often terminates with a missing requested output file or a blank OHLCV conversion crash. The Bayesian run is more expensive, but it keeps posterior-weighted guardrails active for file creation, cache usage, blank-row filtering, and output-format validation, which turns more attempts into valid artifacts.

\section{Case Study: How Skills Evolve}
\label{sec:case_study}

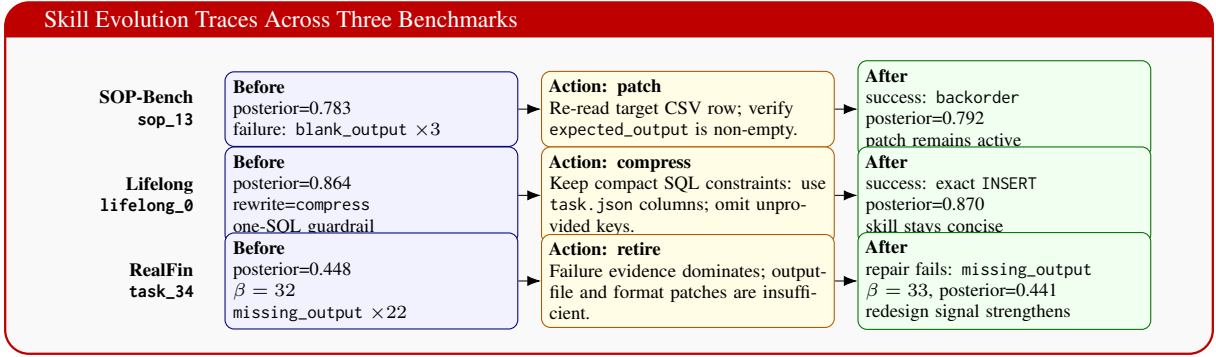
\begin{figure*}[t]
\centering
\small
\begin{tcolorbox}[colback=gray!5!white, colframe=red!75!black,
title=Skill Evolution Traces Across Three Benchmarks, boxrule=0.3mm, width=\textwidth, arc=3mm, auto outer arc=true]
\centering
\begin{tikzpicture}[
  node distance=0.30cm and 0.30cm,
  box/.style={draw, rounded corners=1.2mm, align=left, text width=0.245\textwidth, inner sep=3pt, font=\scriptsize},
  bench/.style={font=\bfseries\scriptsize, align=right, text width=0.105\textwidth},
  arrow/.style={-{Latex[length=1.8mm]}, line width=0.35pt},
  before/.style={box, fill=blue!5, draw=blue!50!black},
  action/.style={box, fill=yellow!12, draw=orange!70!black},
  after/.style={box, fill=green!6, draw=green!45!black}
]
\node[bench] (soplabel) {SOP-Bench\\\texttt{sop\_13}};
\node[before, right=of soplabel] (sopbefore) {\textbf{Before}\\posterior=0.783\\failure: \texttt{blank\_output} \(\times 3\)};
\node[action, right=of sopbefore] (sopaction) {\textbf{Action: patch}\\Re-read target CSV row; verify \texttt{expected\_output} is non-empty.};
\node[after, right=of sopaction] (sopafter) {\textbf{After}\\success: \texttt{backorder}\\posterior=0.792\\patch remains active};
\draw[arrow] (sopbefore) -- (sopaction);
\draw[arrow] (sopaction) -- (sopafter);

\node[bench, below=0.42cm of soplabel] (lifelabel) {Lifelong\\\texttt{lifelong\_0}};
\node[before, right=of lifelabel] (lifebefore) {\textbf{Before}\\posterior=0.864\\rewrite=\texttt{compress}\\one-SQL guardrail};
\node[action, right=of lifebefore] (lifeaction) {\textbf{Action: compress}\\Keep compact SQL constraints: use \texttt{task.json} columns; omit unprovided keys.};
\node[after, right=of lifeaction] (lifeafter) {\textbf{After}\\success: exact \texttt{INSERT}\\posterior=0.870\\skill stays concise};
\draw[arrow] (lifebefore) -- (lifeaction);
\draw[arrow] (lifeaction) -- (lifeafter);

\node[bench, below=0.42cm of lifelabel] (realfinlabel) {RealFin\\\texttt{task\_34}};
\node[before, right=of realfinlabel] (realfinbefore) {\textbf{Before}\\posterior=0.448\\\(\beta=32\)\\\texttt{missing\_output} \(\times 22\)};
\node[action, right=of realfinbefore] (realfinaction) {\textbf{Action: retire}\\Failure evidence dominates; output-file and format patches are insufficient.};
\node[after, right=of realfinaction] (realfinafter) {\textbf{After}\\repair fails: \texttt{missing\_output}\\\(\beta=33\), posterior=0.441\\redesign signal strengthens};
\draw[arrow] (realfinbefore) -- (realfinaction);
\draw[arrow] (realfinaction) -- (realfinafter);
\end{tikzpicture}
\end{tcolorbox}
\caption{Representative skill-evolution traces. SOP-Bench shows a recurring failure mode becoming a concrete patch, Lifelong AgentBench shows stable evidence leading to compact skill context, and RealFin-Bench shows repeated output-file failures becoming an explicit retire/redesign signal.}
\label{fig:skill_evolution_cases}
\end{figure*}

The skill-evolution records make the Bayesian update process inspectable at the text level. Full mode records before/after pairs for each task, while incremental mode records pairs for failed baseline tasks selected for repair. In the RealFin flash setting, full mode records 40 pairs and incremental mode records 22 repair-attempt pairs; the SOP/Lifelong incremental runs likewise record targeted repairs. Each pair contains a posterior audit, model-facing skill context, a belief snapshot, and the verified task result. Figure~\ref{fig:skill_evolution_cases} visualizes one representative trace from each benchmark to show three outcomes: patching a repeated failure, compressing a stable skill, and retiring or redesigning a persistently unreliable skill.

\paragraph{Benchmark-specific evolution.}
The three traces cover three posterior-guided actions: patch, compress, and retire/redesign. In SOP-Bench, three verified blank-output failures promote a guardrail into the model-facing skill context, and the repaired task succeeds with the raw category \texttt{backorder}. In Lifelong AgentBench, the posterior is already high, the task succeeds with an exact SQL statement, and the policy keeps the skill compact instead of adding a long rewrite. In RealFin-Bench, \texttt{task\_34\_etf\_constituent\_arbitrage} has 56 prior observations and a retire decision because missing output files dominate the failure evidence; after the failed repair, the same failure mode is preserved and the posterior falls further, strengthening the redesign signal.

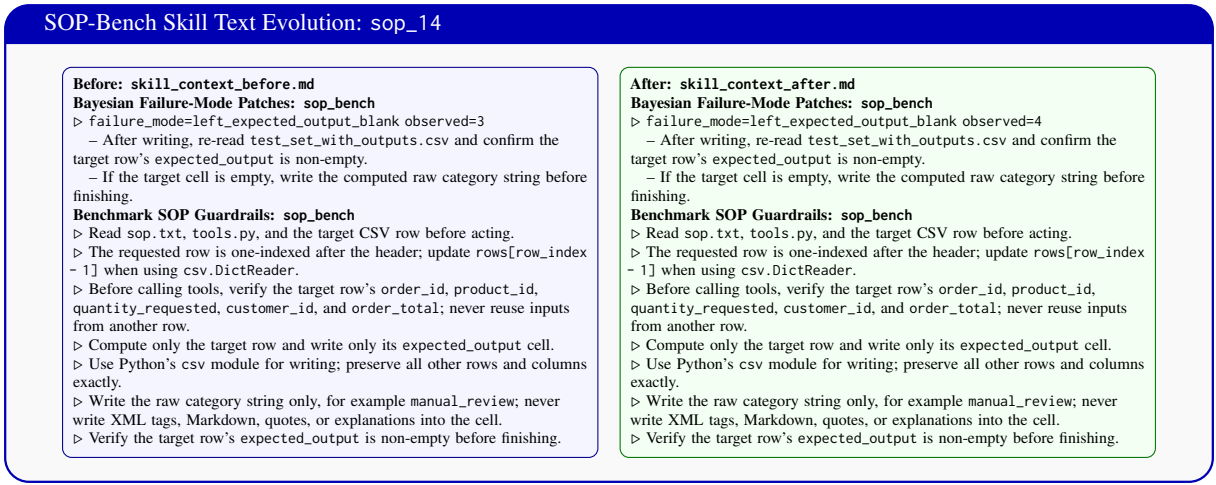
\begin{figure*}[p]
\centering
\small
\begin{tcolorbox}[colback=gray!5!white, colframe=blue!70!black,
title=SOP-Bench Skill Text Evolution: \texttt{sop\_14}, boxrule=0.3mm, width=\textwidth, arc=3mm, auto outer arc=true]
\centering
\begin{tikzpicture}[
  node distance=0.28cm,
  skill/.style={draw, rounded corners=1.2mm, align=left, text width=0.455\textwidth, inner sep=4pt, font=\tiny},
  before/.style={skill, fill=blue!4, draw=blue!55!black},
  after/.style={skill, fill=green!5, draw=green!45!black}
]
\node[before] (sopbefore) {
\textbf{Before: \texttt{skill\_context\_before.md}}\\
\textbf{Bayesian Failure-Mode Patches: \texttt{sop\_bench}}\\
\(\triangleright\) \texttt{failure\_mode=left\_expected\_output\_blank observed=3}\\
\quad -- After writing, re-read \texttt{test\_set\_with\_outputs.csv} and confirm the target row's \texttt{expected\_output} is non-empty.\\
\quad -- If the target cell is empty, write the computed raw category string before finishing.\\
\textbf{Benchmark SOP Guardrails: \texttt{sop\_bench}}\\
\(\triangleright\) Read \texttt{sop.txt}, \texttt{tools.py}, and the target CSV row before acting.\\
\(\triangleright\) The requested row is one-indexed after the header; update \texttt{rows[row\_index - 1]} when using \texttt{csv.DictReader}.\\
\(\triangleright\) Before calling tools, verify the target row's \texttt{order\_id}, \texttt{product\_id}, \texttt{quantity\_requested}, \texttt{customer\_id}, and \texttt{order\_total}; never reuse inputs from another row.\\
\(\triangleright\) Compute only the target row and write only its \texttt{expected\_output} cell.\\
\(\triangleright\) Use Python's \texttt{csv} module for writing; preserve all other rows and columns exactly.\\
\(\triangleright\) Write the raw category string only, for example \texttt{manual\_review}; never write XML tags, Markdown, quotes, or explanations into the cell.\\
\(\triangleright\) Verify the target row's \texttt{expected\_output} is non-empty before finishing.
};
\node[after, right=of sopbefore] (sopafter) {
\textbf{After: \texttt{skill\_context\_after.md}}\\
\textbf{Bayesian Failure-Mode Patches: \texttt{sop\_bench}}\\
\(\triangleright\) \texttt{failure\_mode=left\_expected\_output\_blank observed=4}\\
\quad -- After writing, re-read \texttt{test\_set\_with\_outputs.csv} and confirm the target row's \texttt{expected\_output} is non-empty.\\
\quad -- If the target cell is empty, write the computed raw category string before finishing.\\
\textbf{Benchmark SOP Guardrails: \texttt{sop\_bench}}\\
\(\triangleright\) Read \texttt{sop.txt}, \texttt{tools.py}, and the target CSV row before acting.\\
\(\triangleright\) The requested row is one-indexed after the header; update \texttt{rows[row\_index - 1]} when using \texttt{csv.DictReader}.\\
\(\triangleright\) Before calling tools, verify the target row's \texttt{order\_id}, \texttt{product\_id}, \texttt{quantity\_requested}, \texttt{customer\_id}, and \texttt{order\_total}; never reuse inputs from another row.\\
\(\triangleright\) Compute only the target row and write only its \texttt{expected\_output} cell.\\
\(\triangleright\) Use Python's \texttt{csv} module for writing; preserve all other rows and columns exactly.\\
\(\triangleright\) Write the raw category string only, for example \texttt{manual\_review}; never write XML tags, Markdown, quotes, or explanations into the cell.\\
\(\triangleright\) Verify the target row's \texttt{expected\_output} is non-empty before finishing.
};
\end{tikzpicture}
\end{tcolorbox}
\caption{Before/after model-facing skill text for SOP-Bench. The evidence count for the recurring blank-output patch increases from \texttt{observed=3} to \texttt{observed=4}, while the executable guardrails remain stable.}
\label{fig:sop_skill_text_before_after}
\end{figure*}

Figures~\ref{fig:sop_skill_text_before_after}--\ref{fig:realfin_skill_text_before_after} present benchmark-specific before/after model-facing skill texts for one representative task from each benchmark. These figures show the exact text injected into the harness, so the reader can inspect whether the update is a narrow executable guardrail, a compact preservation decision, or a broader failure-mode patch.

\begin{figure*}[p]
\centering
\small
\begin{tcolorbox}[colback=gray!5!white, colframe=blue!70!black,
title=Lifelong AgentBench Skill Text Evolution: \texttt{lifelong\_12}, boxrule=0.3mm, width=\textwidth, arc=3mm, auto outer arc=true]
\centering
\begin{tikzpicture}[
  node distance=0.28cm,
  skill/.style={draw, rounded corners=1.2mm, align=left, text width=0.455\textwidth, inner sep=4pt, font=\tiny},
  before/.style={skill, fill=blue!4, draw=blue!55!black},
  after/.style={skill, fill=green!5, draw=green!45!black}
]
\node[before] (lifebefore) {
\textbf{Before: \texttt{skill\_context\_before.md}}\\
\textbf{Benchmark SOP Guardrails: \texttt{lifelong\_agentbench}}\\
\(\triangleright\) Read \texttt{task.json} in the current workspace; do not inspect sibling benchmark runs.\\
\(\triangleright\) Write exactly one SQL statement to \texttt{answer.sql}; no Markdown and no explanation.\\
\(\triangleright\) Use only columns present in \texttt{task.json} unless the instruction explicitly asks for a new value in an existing column.\\
\(\triangleright\) For \texttt{INSERT} statements, do not include id or primary-key columns unless the instruction explicitly provides their values.\\
\(\triangleright\) For mutation tasks, write executable SQL that reproduces the expected table state.\\
\(\triangleright\) If SQL ranking is needed, express ranking inside a subquery and keep the final output to one SQL statement.
};
\node[after, right=of lifebefore] (lifeafter) {
\textbf{After: \texttt{skill\_context\_after.md}}\\
\textbf{Bayesian Failure-Mode Patches: \texttt{lifelong\_agentbench}}\\
\(\triangleright\) \texttt{failure\_mode=wrote\_transcript\_instead\_of\_sql}\\
\quad \texttt{after\_workspace\_confusion observed=2}\\
\quad -- Write exactly one executable SQL statement to \texttt{answer.sql}; do not write transcript text, tool logs, Markdown, or explanations.\\
\quad -- Read only the current task workspace and avoid copying content from sibling benchmark runs.\\
\textbf{Benchmark SOP Guardrails: \texttt{lifelong\_agentbench}}\\
\(\triangleright\) Read \texttt{task.json} in the current workspace; do not inspect sibling benchmark runs.\\
\(\triangleright\) Write exactly one SQL statement to \texttt{answer.sql}; no Markdown and no explanation.\\
\(\triangleright\) Use only columns present in \texttt{task.json} unless the instruction explicitly asks for a new value in an existing column.\\
\(\triangleright\) For \texttt{INSERT} statements, do not include id or primary-key columns unless the instruction explicitly provides their values.\\
\(\triangleright\) For mutation tasks, write executable SQL that reproduces the expected table state.\\
\(\triangleright\) If SQL ranking is needed, express ranking inside a subquery and keep the final output to one SQL statement.
};
\end{tikzpicture}
\end{tcolorbox}
\caption{Before/after model-facing skill text for Lifelong AgentBench. The after-state adds a targeted Bayesian failure-mode patch for transcript-like answers caused by workspace confusion, while preserving the compact SQL guardrails.}
\label{fig:lifelong_skill_text_before_after}
\end{figure*}

\begin{figure*}[p]
\centering
\small
\begin{tcolorbox}[colback=gray!5!white, colframe=blue!70!black,
title=RealFin-Bench Skill Text Evolution: \texttt{task\_16\_pe\_bollinger\_reversal}, boxrule=0.3mm, width=\textwidth, arc=3mm, auto outer arc=true]
\centering
\begin{tikzpicture}[
  node distance=0.28cm,
  skill/.style={draw, rounded corners=1.2mm, align=left, text width=0.455\textwidth, inner sep=4pt, font=\tiny},
  before/.style={skill, fill=blue!4, draw=blue!55!black},
  after/.style={skill, fill=green!5, draw=green!45!black}
]
\node[before] (realbefore) {
\textbf{Before: \texttt{skill\_context\_before.md}}\\
\textbf{Bayesian Failure-Mode Patches: \texttt{realfin\_benchmark}}\\
\(\triangleright\) \texttt{failure\_mode=invalid\_realfin\_output\_format observed=2}\\
\quad -- Match the prompt's output format exactly: headers, comma-separated columns, code format, numeric precision, and sort order.\\
\quad -- For stock-code outputs, strip cache prefixes like \texttt{sz.} and \texttt{sh.} unless the task explicitly requests prefixed codes.\\
\quad -- Re-read the output file and validate it against the task's automated format constraints before finishing.\\
\textbf{Benchmark SOP Guardrails: \texttt{realfin\_benchmark}}\\
\(\triangleright\) Read \texttt{task.json} and \texttt{realfin\_cache\_manifest.json} in the current workspace before calculating.\\
\(\triangleright\) Use the local \texttt{api\_cache} symlink for market data; do not call EastMoney historical endpoints such as \texttt{push2his.eastmoney.com}.\\
\(\triangleright\) Create exactly the requested output file in the workspace; do not wrap the file content in Markdown.\\
\(\triangleright\) Map ChiNext code \texttt{300XXX} to baostock CSV\\
\quad \texttt{api\_cache/baostock/daily\_qfq\_20230101\_20260331/}\\
\quad \texttt{sz.300XXX.csv}.\\
\(\triangleright\) When writing stock codes to output files, strip cache market prefixes unless explicitly requested: use \texttt{300531}, not \texttt{sz.300531}.\\
\(\triangleright\) Use auxiliary baostock cache for indexes such as \texttt{sh.000001} and \texttt{sz.399006}.\\
\(\triangleright\) Use Tencent ETF cache files for ETF symbols such as \texttt{sz159642} or \texttt{sh511010}.\\
\(\triangleright\) When a task asks for indicators or constraints, compute them from cached OHLCV data and keep the output format aligned with the prompt.\\
\(\triangleright\) Filter cached rows to valid trading rows with non-empty numeric OHLCV fields; skip blank rows instead of crashing numeric conversion.
};
\node[after, right=of realbefore] (realafter) {
\textbf{After: \texttt{skill\_context\_after.md}}\\
\textbf{Bayesian Failure-Mode Patches: \texttt{realfin\_benchmark}}\\
\(\triangleright\) \texttt{failure\_mode=invalid\_realfin\_output\_format observed=2}\\
\quad -- Match the prompt's output format exactly: headers, comma-separated columns, code format, numeric precision, and sort order.\\
\quad -- For stock-code outputs, strip cache prefixes like \texttt{sz.} and \texttt{sh.} unless the task explicitly requests prefixed codes.\\
\quad -- Re-read the output file and validate it against the task's automated format constraints before finishing.\\
\(\triangleright\) \texttt{failure\_mode=missing\_requested\_output\_file observed=2}\\
\quad -- Before finishing, list the task's requested \texttt{.txt} output file and verify it exists in the workspace.\\
\quad -- If calculations find no qualifying symbols, still create the requested file with the task-accepted empty-result wording or header.\\
\textbf{Benchmark SOP Guardrails: \texttt{realfin\_benchmark}}\\
\(\triangleright\) Read \texttt{task.json} and \texttt{realfin\_cache\_manifest.json} in the current workspace before calculating.\\
\(\triangleright\) Use the local \texttt{api\_cache} symlink for market data; do not call EastMoney historical endpoints such as \texttt{push2his.eastmoney.com}.\\
\(\triangleright\) Create exactly the requested output file in the workspace; do not wrap the file content in Markdown.\\
\(\triangleright\) Map ChiNext code \texttt{300XXX} to baostock CSV\\
\quad \texttt{api\_cache/baostock/daily\_qfq\_20230101\_20260331/}\\
\quad \texttt{sz.300XXX.csv}.\\
\(\triangleright\) When writing stock codes to output files, strip cache market prefixes unless explicitly requested: use \texttt{300531}, not \texttt{sz.300531}.\\
\(\triangleright\) Use auxiliary baostock cache for indexes such as \texttt{sh.000001} and \texttt{sz.399006}.\\
\(\triangleright\) Use Tencent ETF cache files for ETF symbols such as \texttt{sz159642} or \texttt{sh511010}.\\
\(\triangleright\) When a task asks for indicators or constraints, compute them from cached OHLCV data and keep the output format aligned with the prompt.\\
\(\triangleright\) Filter cached rows to valid trading rows with non-empty numeric OHLCV fields; skip blank rows instead of crashing numeric conversion.
};
\end{tikzpicture}
\end{tcolorbox}
\caption{Before/after model-facing skill text for RealFin-Bench. The after-state adds a missing-output-file patch, making file creation and empty-result handling explicit in addition to the existing output-format patch and data-cache guardrails.}
\label{fig:realfin_skill_text_before_after}
\end{figure*}

\paragraph{Interpretation.}
These cases provide interpretability evidence by preserving a traceable chain from verifier outcome to evidence features, from evidence features to posterior audit, and from posterior audit to model-facing skill edits. They are qualitative traces showing how successful repairs, stable compressions, and negative redesign signals remain visible to the developer under an explicit Bayesian evidence frame.

\section{Administrative Notes}
\label{sec:appendix_admin}

\subsection{Limitations}

Backend and model coverage remain limited. The main comparison centers on GenericAgent, while the backend ablation adds the native Bayesian-Agent backend, MiniSWEAgent, and Claude Code. This supports cross-harness portability but does not establish broad plug-and-play generality across arbitrary harness/model pairs. 
The default backend is a factorized categorical evidence model with Laplace smoothing rather than full Bayesian structure learning or model selection. 

\subsection{Ethical Considerations}

This work studies harness-side reliability mechanisms using benchmark tasks and existing experiment records; no human-subject data are collected. Skill repair can reduce repeated operational failures, but it can also make an agent more persistent. Bayesian-Agent therefore preserves posterior audits and exposes failure-mode patches for inspection. Deployment still requires permission checks, logging, and human oversight because risks inherited from the base model, harness, tools, and benchmark data remain.

\subsection{Information About Use of AI Assistants}

AI-assisted tools were used for proofreading and language refinement. The authors remain responsible for the manuscript's content, claims, and verification.

\end{document}